%% file: main.tex
\documentclass{article} % For LaTeX2e
\usepackage{iclr2027_conference,times}
\iclrfinalcopy

\input{math_commands.tex}

\usepackage{hyperref}
\usepackage{url}
\usepackage[utf8]{inputenc}
\usepackage{amssymb}
\usepackage{blindtext}
\usepackage{listings}

\usepackage{xcolor}
\usepackage{tcolorbox}

\usepackage{pythonhighlight}
\usepackage{tabularx}
\usepackage{cleveref}
\usepackage{array}
\usepackage{booktabs}
\usepackage{xcolor}
\usepackage{colortbl}
\usepackage{amsmath}
\usepackage{amsthm}
\usepackage{float}
\usepackage{subcaption}
\usepackage{tikz}
\usetikzlibrary{positioning, arrows.meta, calc, fit, backgrounds}

\definecolor{codegreen}{rgb}{0,0.6,0}
\definecolor{codegray}{rgb}{0.5,0.5,0.5}
\definecolor{codepurple}{rgb}{0.58,0,0.82}
\definecolor{backcolour}{rgb}{0.95,0.95,0.92}

\newcommand{\SE}[1]{\,{\scriptsize\textcolor{gray!70}{$\pm$#1}}}

\title{Mixture of GFlowNets}

\author{
\textbf{Omar G. Younis}$^{\star,1}$ \quad
\textbf{Joseph D. Viviano}$^{\star,1}$ \quad
\textbf{Chirayu Anant Haryan}$^{3}$ \\
\textbf{MD Vasimuddin}$^{3}$ \quad
\textbf{Sanchit Misra}$^{3}$ \quad
\textbf{Yoshua Bengio}$^{1,2}$ \\[0.6em]
$^{1}$Mila -- Qu\'ebec AI Institute \quad
$^{2}$Universit\'e de Montr\'eal \quad
$^{3}$Intel \\[0.3em]
\parbox{0.9\textwidth}{\small
$^{\star}$Equal contribution; ordered by geographic proximity to the Duomo di Milano at the time of publication. Co-first authors may list their names in either order on their CVs.} \vspace{2mm} \\
\texttt{omar.g.younis@gmail.com}, \;
\texttt{joseph@viviano.ca}, \\
\texttt{\{chirayu.anant.haryan, sanchit.misra\}@intel.com} \\[0.5em]
}

\begin{document}

\maketitle

\begin{abstract}%
    Generative Flow Networks (GFlowNets) learn a policy that samples compositional objects in proportion to an unnormalized reward $R(x)$. This makes them useful wherever many diverse, high-reward candidates are needed and oracle queries are expensive, as in AI for Science. In practice, a single GFlowNet often fails to cover all modes of a complex distribution, and the obvious ways to scale compute saturate: larger batches and data-parallel gradient averaging draw every trajectory from one, possibly biased, policy, so extra rollouts rediscover known modes instead of finding new ones. We explore whether this mode-bias problem (mode collapse in the extreme) can be addressed with a population of asynchronously trained policies which divide the state space between them.
    \textbf{Mixture of GFlowNets (MoG)} replaces the single forward policy with a \textit{community} of $K$ independent GFlowNets coupled through one jointly trained soft classifier, which assigns each member its own region of the state space through reward shaping. Mixing the members under their learned partition functions provably recovers the target $p(x)\propto R(x)$. \textbf{MoG exchange (MoGX)} also lets the classifier route sampled trajectories: a trajectory the classifier assigns to another member is sent to that member and trained on there, with no importance sampling.
    We evaluate both methods by keeping the environment-interaction budget fixed while scaling the community size, turning parallel \emph{compute} into mode \emph{coverage} without extra oracle calls. On HyperGrid, across six reward landscapes and communities of up to $K{=}64$, MoG and MoGX keep finding new modes as $K$ grows, while data-parallel training saturates. On a genomic ideotype-design task, MoGX covers $39\%$ of the $1024$ optima at $K{=}64$, against $31\%$ for independent agents and $20\%$ for data-parallel training. MoG also has a lower per-iteration synchronization cost than data-parallel training.
\end{abstract}

\section{Introduction}
\label{sec:introduction}
\vspace{-2mm}

Many \emph{AI for Science} problems, such as drug design, materials discovery and biological sequence engineering, require searching huge combinatorial spaces for many distinct, high-reward objects. Here \emph{sample diversity} is the goal, and the dominant cost is the environment query rather than gradient compute, so the number of sampled trajectories should be minimized. Generative Flow Networks (GFlowNets) fit this regime \citep{bengio2021flow,jain2022biological,bengio2023gflownet,pan2024pretraining}. They amortize sampling from an unnormalized reward $R(x)$, so the trained policy samples $x \propto R(x)$ and spreads over the high-reward regions instead of collapsing onto one, as \emph{return-maximizing} RL tends to \citep{malkin2022trajectory,madan2023learning}.\footnote{Entropy-regularized RL and samplers for unnormalized densities are closely related; see \cref{app:related}.}

In practice, trained GFlowNets often miss modes and concentrate on a subset of the high-reward regions; \citet{kim2025boosted_gfn} report this on synthetic benchmarks and on peptide design. Two lines of work try to fix this. The first adds exploration to a single policy: intermediate rewards, intrinsic bonuses, Thompson sampling, local search, replay prioritization, or exploratory ``sibling'' networks \citep{pan2023gafn,kim2023local,rector2023tsgfn, shen2023towards, schaul2016prioritized, vemgal2023empirical, madan2025sibling}. These raise trajectory diversity, but all exploration happens around one learned sampler, and a biased sampler limits what can be reached at any iteration. The second uses several GFlowNets. \citet{kim2025boosted_gfn} fit a chain of GFlowNets, each targeting the reward its predecessors left uncovered, which helps coverage but is serial, and therefore cannot leverage parallel compute. In contrast, \citet{ikram2024evolution} (EGFN) evolve a parallel population of GFlowNet agents by weight mutation and distill the population's best trajectories into one gradient-trained sampler through a prioritized replay buffer. The population's search is uncoordinated, and so it serves as an exploration front for that one sampler.

Adding parallel rollouts during training of a single policy gives \emph{saturating} returns. \citet{obando2023small} found that \emph{smaller} batches improve value-based deep RL, and \citet{singla2024sapg} show that on-policy policy-gradient training plateaus at tens of thousands of parallel environments, because extra samples from one policy add diminishing trajecotory diversity. SAPG overcomes this plateau by dividing the environments among several sub-policies, but the split helps only when each sub-policy also learns, off-policy, from the data the others collect. What scales is data that a single policy would not produce. GFlowNets saturate for the same reason: Every trajectory in a batch comes from the same forward policy $P_F^{\pi_\theta^i}$, so extra trajectories land in modes the policy already knows and carry no additional signal about missing modes. If the policy's \emph{reachable set} $\mathcal{R}^{\pi_\theta^i}(s_0)$ at iteration $i$ is a strict subset of the terminal space $\mathcal{X}$, no batch size closes the gap. This is known as the \emph{mode-bias} problem (which leads to \emph{mode collapse} in the extreme).

\vspace{-2mm}
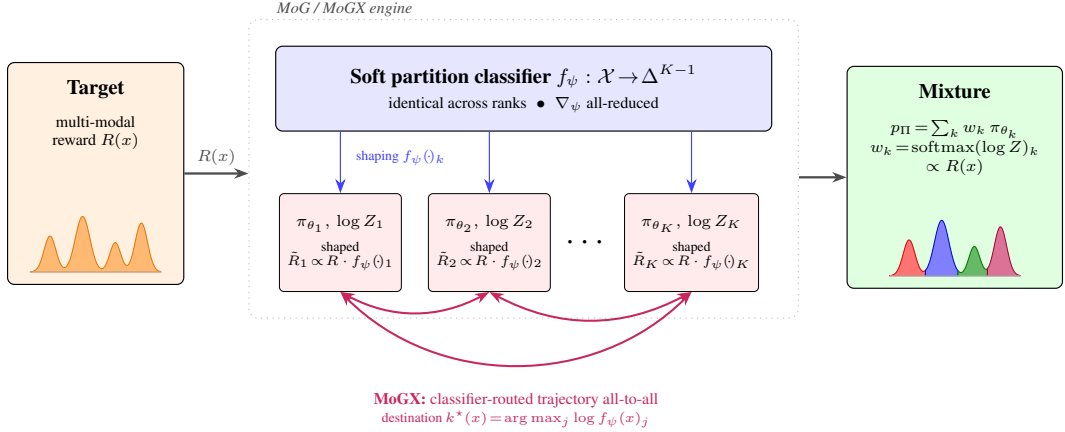
\begin{figure}[h]
    \centering
    \resizebox{\linewidth}{!}{%
    \begin{tikzpicture}[
        font=\small,
        >={Stealth[length=2.2mm,width=1.6mm]},
        target/.style={draw, rounded corners=3pt, fill=orange!12, line width=0.5pt, align=center},
        clf/.style={draw, rounded corners=3pt, fill=blue!10, line width=0.5pt, align=center},
        agent/.style={draw, rounded corners=2pt, fill=red!8, line width=0.4pt, align=center,
                      font=\scriptsize, minimum width=18mm, minimum height=15mm},
        mix/.style={draw, rounded corners=3pt, fill=green!12, line width=0.5pt, align=center},
        ph/.style={draw, dashed, line width=0.3pt, font=\tiny\itshape, text=gray!55!black,
                   align=center, fill=white, inner sep=2pt},
        rshape/.style={->, line width=0.5pt, draw=blue!75},
        sync/.style={->, line width=0.5pt, draw=blue!75, dashed},
        exch/.style={<->, line width=0.9pt, draw=purple!85},
        bigarrow/.style={->, line width=1pt, draw=black!70},
    ]
    % --- Target reward (left) ---
    \node[target, minimum width=27mm, minimum height=34mm] (target) at (0,0) {};
    \node[anchor=north] at ([yshift=-1.5mm]target.north) {\textbf{Target}};
    \node[anchor=north, font=\scriptsize, align=center] at ([yshift=-7mm]target.north)
        {multi-modal\\reward $R(x)$};
    % Target reward heatmap: single-color multi-modal landscape
    \begin{scope}[shift={([yshift=2mm]target.south)}]
      \draw[fill=orange!55, draw=orange!90!black, line width=0.3pt]
        plot[domain=-1:1, samples=70, smooth]
          (\x, {0.55*exp(-(\x+0.7)*(\x+0.7)*60)
              + 0.85*exp(-(\x+0.2)*(\x+0.2)*40)
              + 0.45*exp(-(\x-0.3)*(\x-0.3)*70)
              + 0.75*exp(-(\x-0.7)*(\x-0.7)*55)})
        -- (1,0) -- (-1,0) -- cycle;
      \draw[gray!55, line width=0.2pt] (-1,0) -- (1,0);
    \end{scope}
    
    % --- Classifier (top, wide) ---
    \node[clf, minimum width=76mm, minimum height=13mm] (clf)
        at ($(target.east)+(52mm,13mm)$) {%
        \textbf{Soft partition classifier $f_\psi:\mathcal{X}\!\to\!\Delta^{K-1}$}\\[1pt]
        {\scriptsize identical across ranks \,\,$\bullet$\,\, $\nabla_\psi$ all-reduced}};
    
    % --- K agents (community) ---
    \node[agent] (a1) at ($(clf.south west)+(10mm,-17mm)$) {%
        $\pi_{\theta_1},\,\log Z_1$\\[1mm]
        \tiny shaped\\[-0.7mm]
        \tiny $\tilde R_1\!\propto\! R\cdot f_\psi(\!\cdot\!)_1$};
    \node[agent, right=4mm of a1] (a2) {%
        $\pi_{\theta_2},\,\log Z_2$\\[1mm]
        \tiny shaped\\[-0.7mm]
        \tiny $\tilde R_2\!\propto\! R\cdot f_\psi(\!\cdot\!)_2$};
    \node[right=1mm of a2, font=\Large] (dots) {$\cdots$};
    \node[agent, right=1mm of dots] (aK) {%
        $\pi_{\theta_K},\,\log Z_K$\\[1mm]
        \tiny shaped\\[-0.7mm]
        \tiny $\tilde R_K\!\propto\! R\cdot f_\psi(\!\cdot\!)_K$};
    
    % --- Mixture (right) ---
    \node[mix, minimum width=33mm, minimum height=34mm] (mix)
        at ($(aK.east)+(30mm,10mm)$) {};
    \node[anchor=north] at ([yshift=-1.5mm]mix.north) {\textbf{Mixture}};
    \node[anchor=north, font=\scriptsize, align=center] at ([yshift=-7mm]mix.north)
        {$p_\Pi\!=\!\sum_k w_k\,\pi_{\theta_k}$\\
         $w_k\!=\!\mathrm{softmax}(\log Z)_k$\\
         $\propto R(x)$};
    % Recovered heatmap: same envelope as target, partitioned into K colored components
    \begin{scope}[shift={([yshift=2mm]mix.south)}]
      % Component k=1 (red): owns the leftmost mode, x in [-1, -0.45]
      \draw[fill=red!55, draw=red!85!black, line width=0.25pt]
        plot[domain=-1:-0.45, samples=22, smooth]
          (\x, {0.55*exp(-(\x+0.7)*(\x+0.7)*60)
              + 0.85*exp(-(\x+0.2)*(\x+0.2)*40)
              + 0.45*exp(-(\x-0.3)*(\x-0.3)*70)
              + 0.75*exp(-(\x-0.7)*(\x-0.7)*55)})
        -- (-0.45,0) -- (-1,0) -- cycle;
      % Component k=2 (blue): owns the second mode, x in [-0.45, 0.05]
      \draw[fill=blue!50, draw=blue!80!black, line width=0.25pt]
        plot[domain=-0.45:0.05, samples=22, smooth]
          (\x, {0.55*exp(-(\x+0.7)*(\x+0.7)*60)
              + 0.85*exp(-(\x+0.2)*(\x+0.2)*40)
              + 0.45*exp(-(\x-0.3)*(\x-0.3)*70)
              + 0.75*exp(-(\x-0.7)*(\x-0.7)*55)})
        -- (0.05,0) -- (-0.45,0) -- cycle;
      % Component k=3 (green): owns the third mode, x in [0.05, 0.5]
      \draw[fill=green!55!black!60, draw=green!45!black, line width=0.25pt]
        plot[domain=0.05:0.5, samples=22, smooth]
          (\x, {0.55*exp(-(\x+0.7)*(\x+0.7)*60)
              + 0.85*exp(-(\x+0.2)*(\x+0.2)*40)
              + 0.45*exp(-(\x-0.3)*(\x-0.3)*70)
              + 0.75*exp(-(\x-0.7)*(\x-0.7)*55)})
        -- (0.5,0) -- (0.05,0) -- cycle;
      % Component k=K (purple): owns the rightmost mode, x in [0.5, 1]
      \draw[fill=purple!55, draw=purple!85!black, line width=0.25pt]
        plot[domain=0.5:1, samples=22, smooth]
          (\x, {0.55*exp(-(\x+0.7)*(\x+0.7)*60)
              + 0.85*exp(-(\x+0.2)*(\x+0.2)*40)
              + 0.45*exp(-(\x-0.3)*(\x-0.3)*70)
              + 0.75*exp(-(\x-0.7)*(\x-0.7)*55)})
        -- (1,0) -- (0.5,0) -- cycle;
      \draw[gray!55, line width=0.2pt] (-1,0) -- (1,0);
    \end{scope}
    
    % --- Engine background grouping ---
    \begin{scope}[on background layer]
    \node[draw=gray!50, dotted, line width=0.5pt, rounded corners=4pt,
          inner sep=4mm, fit=(clf)(a1)(aK)] (engine) {};
    \end{scope}
    \node[anchor=south west, font=\scriptsize\itshape, text=gray!60!black, inner sep=1pt]
        at ([xshift=4mm, yshift=0mm]engine.north west) {MoG / MoGX engine};
    
    % ===== Edges =====
    % Target -> engine (single thick arrow carrying R(x))
    \draw[bigarrow] (target.east) -- (target.east -| engine.west)
        node[midway, above, font=\scriptsize, text=black!70] {$R(x)$};
    
    % Classifier -> each agent (reward shaping, blue solid)
    \draw[rshape] (clf.south -| a1.north) -- (a1.north);
    \draw[rshape] (clf.south -| a2.north) -- (a2.north);
    \draw[rshape] (clf.south -| aK.north) -- (aK.north);
    \node[font=\tiny, text=blue!75, anchor=west, inner sep=1pt]
        at ($(clf.south -| a1.north) + (2mm,-4mm)$) {shaping $f_\psi(\!\cdot\!)_k$};
    
    % Sync back-arrow (gradient all-reduce on f_psi); label folded into
    % the classifier subtitle ("only $\nabla_\psi$ all-reduced") to avoid clutter.
    % \draw[sync] ($(aK.north)+(-2mm,1mm)$)
    %     .. controls +(up:7mm) and +(right:7mm) .. ($(clf.east)+(-3mm,-1mm)$);
    
    % MoGX exchange (purple curved arrows under the agents)
    \draw[exch] (a1.south) to[bend right=28] (a2.south);
    \draw[exch] (a2.south) to[bend right=28] (aK.south);
    \draw[exch] (a1.south) to[bend right=45] (aK.south);
    \node[font=\scriptsize, text=purple!85, align=center, anchor=north, inner sep=1pt]
        at ($(a1.south)!0.5!(aK.south)+(0,-15mm)$)
        {\textbf{MoGX:} classifier-routed trajectory all-to-all\\
         {\tiny destination $k^\star(x)\!=\!\arg\max_j \log f_\psi(x)_j$ }};
    
    % Engine -> mixture
    \draw[bigarrow] (mix.west -| engine.east) -- (mix.west);
    \end{tikzpicture}%
    }
    \caption{\textbf{Mixture of GFlowNets (MoG) and MoG exchange (MoGX) at a
    glance.} A community of $K$ GFlowNets is trained in parallel against a
    target $R(x)$, coupled to one jointly trained soft
    classifier $f_\psi$ that softly partitions the state space. Each GFlowNet trains
    on the \emph{shaped} reward (blue
    arrows). The classifier is the \emph{only} object whose gradients cross rank boundaries.
    \textbf{MoGX} routes part of the on-policy trajectory to other ranks based on the classifier. Post-training, the
    $Z$-weighted mixture provably recovers $p(x)\!\propto\!R(x)$.}

    \label{fig:pipeline}
    \vspace{-2mm}

\end{figure}

Existing work scales GFlowNet training asynchronously without addressing this bias. Asynchronous trajectory balance \citep{bartoldson2025trajectory} spreads off-policy searchers across nodes that feed a shared buffer, but the searchers are snapshots of one policy and share its reachable set. Embarrassingly Parallel GFlowNets \citep{da2024embarrassingly} train one local GFlowNet per factor of a product reward $R(\cdot) \propto R_1(\cdot) \cdots R_N(\cdot)$ to distribute reward evaluation, not to cover a shared target better. Distributed RL \citep{espeholt2018impala,horgan2018distributed} and population-based training \citep{jaderberg2017population} parallelize agents without encouraging specialization (\cref{app:related}).

We attack mode bias with a population of policies that divide the state space between them. \textbf{Mixture of GFlowNets (MoG)} replaces the single forward policy with a community of $K$ GFlowNets $\Pi_\theta=\{\pi_{\theta_1},\ldots,\pi_{\theta_K}\}$ trained in parallel and coupled to one jointly trained soft classifier $f_\psi:\mathcal{X}\!\to\!\Delta^{K-1}$ that partitions $\mathcal{X}$. Component $k$ trains on the shaped reward $\log\tilde R_k(x)=\log R(x)+\log f_\psi(x)_k$, and mixing the components under their learned partition functions recovers $p(x)\propto R(x)$ exactly. Only the small classifier's gradients are all-reduced across ranks, bounding communication costs, and the classifier pushes each GFlowNet toward its own subregion. This removes both the parameter coupling that limits data-parallel training and the redundancy that limits independent training. \textbf{MoG exchange (MoGX)} goes further and sends each on-policy trajectory to the rank the classifier assigns it to, through an MPI all-to-all keyed on $\arg\max_j \log f_\psi(x)_j$. GFlowNet losses accept off-policy data, so the exchange needs no importance correction. \Cref{fig:pipeline} shows the full pipeline.

% \paragraph{Contributions.}
% \begin{itemize}
%     \item A formalization of the \textit{mode bias} problem that explains
%       why single-policy training, na\"ive batch scaling, and data-parallel
%       training all saturate for GFlowNets, with direct consequences for
%       oracle-budget-constrained AI for Science.
%     \item \textbf{MoG}: a community of $K$ GFlowNets coupled to a jointly
%       trained soft classifier that partitions the state space via reward
%       shaping, while provably preserving the mixture target
%       $p(x)\propto R(x)$.
%     \item \textbf{MoGX}: classifier-driven trajectory exchange that supplies
%       each component with in-subregion samples from the first iteration,
%       accelerating concentration of the shaped loss.
%     \item A distributed stack for population-based GFlowNet training at HPC
%       scale (remote prioritized replay buffers, a global mode registry) that
%       makes iso-budget comparisons across community sizes meaningful and
%       reproducible.
%     \item Evidence on HyperGrid across six reward landscapes (non-tiered:
%       \emph{original}, \emph{cosine}, \emph{sparse}; tiered/compositional:
%       \emph{bitwise\_xor}, \emph{multiplicative\_coprime},
%       \emph{conditional\_multiscale}) that MoG and MoGX keep discovering modes
%       as $K$ scales from $1$ to $64$, while DDP saturates (consistent with
%       \citealp{singla2024sapg}) and independent agents plateau through
%       redundancy. MoGX improves over MoG most on the structurally richest
%       rewards.
% \end{itemize}

\section{Background}
\label{sec:background}
\vspace{-2mm}

\paragraph{GFlowNets.}
Let $\mathcal{G} = (\mathcal{S}, \mathcal{A})$ be a finite directed acyclic
graph with a unique source $s_0$ and terminal states
$\mathcal{X} \subseteq \mathcal{S}$ carrying a non-negative reward
$R : \mathcal{X} \to \mathbb{R}_{\geq 0}$. A GFlowNet
\citep{bengio2021flow,bengio2023gflownet} parameterizes a forward policy
$P_F(s' \mid s)$ and a backward policy $P_B(s \mid s')$, trained so that the marginal
over terminal states induced by following $P_F$ from $s_0$ satisfies
$p_\theta(x) \propto R(x)$. The trajectory balance (TB) objective
\citep{malkin2022trajectory} enforces, for every complete trajectory
$\tau = (s_0, \ldots, s_T = x)$,
\[
Z \cdot \prod_{t=0}^{T-1} P_F(s_{t+1} \mid s_t)
\;=\;
R(x) \cdot \prod_{t=1}^{T} P_B(s_{t-1} \mid s_t),
\]
with $Z > 0$ a learnable estimate of $\sum_{x} R(x)$. Detailed balance
\citep{bengio2023gflownet} and sub-trajectory balance \citep{madan2023learning}
give alternative decompositions of the same constraint; our framework does
not depend on the choice of training objective.

\paragraph{Single-learner coverage and the mode-bias problem.}
The guarantee $p_\theta(x) \propto R(x)$ holds at the global optimum of the balance objective, which requires the balance conditions on \emph{every} trajectory. However, training only enforces them on trajectories the behavior distribution visits. On multi-modal landscapes, that distribution often concentrates on a few high-reward regions, so the conditions can be badly violated on undiscovered modes while the empirical loss looks converged \citep{kim2025boosted_gfn}. The remedies in \cref{sec:introduction} change how a single learner explores; some of them sample off-policy, but in all of them the behavior distribution is generated by, or seeded from, one set of learned parameters. The learner's current beliefs alone decide which regions receive gradient signal. We make this precise with the
\emph{$(\epsilon,T)$-reachable set} at iteration $i$,
\[
\mathcal{R}_{\epsilon,T}^{\pi_\theta^{(i)}}(s_0)
\;:=\;
\bigl\{\, x \in \mathcal{X} :
\Pr\bigl[s_0 \rightsquigarrow x \text{ in} \le T \text{ steps under } \pi_\theta^{(i)}\bigr] \ge \epsilon \,\bigr\},
\]
where $\pi_\theta^{(i)}$ is the forward policy after $i$ training iterations. A softmax policy gives every $x$ positive probability, so coverage failure is about non-negligible probability under a realistic budget: $\mathcal{R}_{\epsilon,T}^{\pi_\theta^{(i)}}(s_0)\subsetneq \mathcal{X}$ whenever the path to some $x$ needs transitions whose product of probabilities falls below $\epsilon$. Larger batches help only marginally. Extra trajectories are drawn i.i.d.\ from the same behavior distribution, so the expected cost of reaching an excluded mode scales as $1/\epsilon$, exponentially in the number of low-probability transitions on its path. On-policy deep RL shows the same saturation under batch scaling \citep{singla2024sapg}.

\paragraph{Decorrelating failures across a population.}
Every policy will miss some modes, so we aim for policies that miss different ones, leaving a mode lost only when all of them miss it. $N$ independently initialized policies $\{\pi_\theta^i\}_{i=1}^{N}$ sample from the union $\bigcup_i \mathcal{R}^{\pi_\theta^i}(s_0)$, which is larger than a single reachable set only if the policies fail \emph{differently}. Independent initialization does not guarantee that, since policies trained on the same reward can still settle on the same modes. So we push them apart explicitly, making their reachable sets overlap as little as possible.

\section{Methods}
\label{sec:methods}
\vspace{-2mm}

\subsection{Mixture Framework}
\label{sec:overview}
\vspace{-2mm}

We train a community $\Pi_\theta=\{\pi_{\theta_1},\ldots,\pi_{\theta_K}\}$ in parallel on a target reward $R:\mathcal{X}\to\mathbb{R}_{\geq 0}$ over a finite DAG $\mathcal{G}=(\mathcal{S},\mathcal{A})$ with source $s_0$. Component $k$ owns forward and backward policies and a learnable log-partition $\log Z_k$ (for DB and SubTB, the source log-flow $\log F^{(k)}(s_0)$ plays the same role). The total budget is fixed: instead of $T$ trajectories from one policy, each component gets $T/K$. The components are coupled through one soft classifier $f_\psi:\mathcal{X}\to\Delta^{K-1}$ trained jointly with them, and component $k$ trains on the \emph{shaped} log-reward
\begin{equation}
\log\tilde R_k(x) \;=\; \log R(x) \;+\; \log f_\psi(x)_k,
\label{eq:shaped-reward}
\end{equation}
so that $\pi_{\theta_k}(x)\propto\tilde R_k(x)$ at convergence. Mixing the components under their learned partition functions,
\begin{equation}
\mathcal{P}_\Pi(\pi_k) \;=\; \frac{Z_k}{\sum_{j=1}^{K} Z_j},
\qquad
p_\Pi(x) \;=\; \sum_{k=1}^{K}\mathcal{P}_\Pi(\pi_k)\,\pi_{\theta_k}(x)
\;\propto\; R(x),
\label{eq:mixture-target}
\end{equation}
recovers the original target.

\paragraph{Proposition (mixture correctness).} \emph{If each component reaches
its shaped target, $\pi_{\theta_k}(x) = \tilde R_k(x)/Z_k$ with
$Z_k=\sum_{x}R(x)f_\psi(x)_k$, and $\sum_k f_\psi(x)_k = 1$ pointwise, then $p_\Pi(x) = R(x)/\sum_{x'}R(x')$.}

\noindent\emph{Proof.}
$p_\Pi(x) = \sum_k \frac{Z_k}{\sum_j Z_j}\cdot\frac{R(x)f_\psi(x)_k}{Z_k}
= \frac{R(x)}{\sum_j Z_j}$, and
$\sum_j Z_j=\sum_x R(x)\sum_j f_\psi(x)_j=\sum_x R(x)$. \qed

The result holds for \emph{any} pointwise simplex assignment; $f_\psi$ only redistributes mass. Since $\tilde R_k$ is supported on $\{x : f_\psi(x)_k > 0\}$, a set the classifier can align with component $k$'s reachable set, each component's convergence assumption can be \emph{easier} to meet than reaching all of $\mathcal{X}$. After training, we form $w_k=\mathrm{softmax}(\log Z)_k$, sample $\pi_k\sim\mathrm{Categorical}(w)$ and then $x\sim\pi_{\theta_k}$; \cref{sec:experiments} checks that this reproduces $p(x)\propto R(x)$ (\cref{fig:mog_mixture_correctness}). We train with trajectory balance, but any objective with a log-partition estimate works.

\subsection{Mixture of GFlowNets (MoG)}
\label{sec:mog}
\vspace{-2mm}

MoG jointly trains $K$ GFlowNets and one soft classifier. The classifier is the \emph{only} object whose gradients are all-reduced across ranks. Each GFlowNet otherwise samples its own data and takes its own updates, which keeps communication minimal and lets MoG train asynchronously.

\paragraph{Partition classifier and reward shaping.}
$f_\psi$ reads the same preprocessed state $\phi(x)$ as the policies and outputs $K$ logits. Its gradients are all-reduced before each step, so its parameters stay identical across ranks and the shaped rewards stay coherent (\cref{app:method-details}). Rank $k$ replaces the log-reward of every terminating state with \cref{eq:shaped-reward} at loss time, so the shaping always reflects the \emph{current} classifier, even for states buffered many iterations ago.

\paragraph{Classifier objective.}
Each rank trains $f_\psi$ by cross-entropy, labeling every on-policy state with its own component index:
\begin{equation}
\mathcal{L}_{f_\psi}^{(k)}(\psi)
\;=\;
-\frac{1}{|\mathcal{B}^{(k)}|}\sum_{x\in\mathcal{B}^{(k)}}
   \log f_\psi(x)_k
\;+\;\beta\,\mathcal{H}\bigl(f_\psi(x)\bigr),
\label{eq:fpsi-loss}
\end{equation}
with $\mathcal{H}$ the softmax entropy and $\beta\geq 0$. All-reduced across ranks, this is supervised classification on the union $\bigcup_k \{(x,k):x\in\mathcal{B}^{(k)}\}$, whose Bayes-optimal solution is $f_\psi(x)_k = P(k\mid x) \propto \pi_{\theta_k}(x)$. As each $\pi_{\theta_k}$ specializes, the classifier sharpens, and \cref{eq:shaped-reward} reinforces that specialization on the next iteration, with \cref{eq:mixture-target} as the fixed point.

\subsection{Trajectory Exchange (MoGX)}
\label{sec:mogx}
\vspace{-2mm}

MoG changes each component's \emph{training signal} but not its \emph{experience collection}: rank $k$ still samples from $\pi_{\theta_k}$ alone, which early in training may rarely visit its assigned subregion. MoGX sends each on-policy trajectory to the component the classifier assigns it to.

\paragraph{Selection and transport.}
After sampling its batch $\mathcal{B}^{(k)}$, rank $k$ computes each terminating state's destination
\begin{equation}
k^\star(x_i) \;=\; \arg\max_{j\in\{1,\ldots,K\}} \log f_\psi(x_i)_j .
\label{eq:dest}
\end{equation}
From the states assigned elsewhere, $\mathcal{E}^{(k)}=\{x_i\in\mathcal{B}^{(k)}:k^\star(x_i)\neq k\}$, it sends the $\min(N,|\mathcal{E}^{(k)}|)$ trajectories with the \emph{smallest} $\log f_\psi(x_i)_k$ (those the classifier is most sure do not belong to $k$) to their destination ranks. $N$ caps the per-step communication volume, and components outside an exchange keep sampling and training independently. TB \citep{malkin2022trajectory}, DB \citep{bengio2023gflownet}, SubTB \citep{madan2023learning} and ZVar \citep{zhang2023robust} reach zero loss only when the balance conditions hold on every trajectory the training distribution can produce. Assuming the model can reach zero loss and the training distribution has full support, every such distribution gives the same global minimizer, so exchanged trajectories need no importance-sampling correction. See \Cref{alg:mogx} for the MoGX algorithm. Steps unique to MoGX are marked $\dagger$, removing them recovers MoG, removing both the exchange step and the reward shaping recovers NoComm, and adding a gradient all-reduce on policy parameters recovers DDP. \Cref{app:method-details} additional and communication details.

\begin{figure}[H]
\hrule\vspace{2pt}\hrule\vspace{4pt}
\textbf{Algorithm 1: One MoGX iteration on training rank $k$.}\hfill
\vspace{2pt}\hrule\vspace{6pt}
\begin{enumerate}\itemsep0pt\parsep0pt
  \item \textbf{Sample.} Roll out $|\mathcal{B}|$ trajectories from
        $\pi_{\theta_k}$.
  \item \textbf{Exchange}. Compute $k^\star(x_i)$ and
        self-log-probs via $f_\psi$; select $n_{\text{exch}}$
        low-self-probability misrouted trajectories per
        \cref{eq:dest}; \texttt{MPI\_Alltoall} across the training
        group; merge received trajectories into the local batch.
  \item \textbf{Store.} Insert the merged batch into the local replay buffer $\mathcal{D}$.
  % \item \textbf{Mix.} On-policy: train on the current batch only.
  %       Otherwise, with $\rho=0.5$: half batch from on-policy, half from replay buffer.
  \item \textbf{Mix.} $\mathcal{B} \leftarrow \mathcal{B}_{\text{on}}^{(1-\rho)B} \cup \mathcal{B}_{\mathcal{D}}^{\rho B}$,
      i.e., a fraction $\rho$ of the batch is sampled from the replay buffer $\mathcal{D}$
      and the rest from the current on-policy data ($\rho=0$: on-policy only).
  \item \textbf{Shape rewards.} Replace $\log R(x)$ by
        $\log\tilde R_k(x)$ via \cref{eq:shaped-reward}, using current
        $f_\psi$.
  \item \textbf{GFlowNet loss.} Compute
        $\mathcal{L}_{\text{GFN}}^{(k)}(\theta_k)$ on the shaped batch;
        do a backward pass and the optimizer step.
%   \item \textbf{Policy gradient sync (DDP only).} All-reduce
%         $\nabla_\theta\mathcal{L}_{\text{GFN}}^{(k)}$ across training
%         ranks. Skipped for MoG and MoGX.
%   \item \textbf{Policy step.} Clip
%         $\|\nabla_\theta\mathcal{L}_{\text{GFN}}^{(k)}\|_2\leq 10$;
%         AdamW step with linear warmup over the first $500$ iterations.
  \item \textbf{Classifier step.} Compute \cref{eq:fpsi-loss};
        backward pass; all-reduce $\nabla_\psi\mathcal{L}_{f_\psi}^{(k)}$
        across training ranks; optimizer step on $f_\psi$.
\end{enumerate}
\vspace{6pt}\hrule\vspace{2pt}\hrule
\label{alg:mogx}
\end{figure}

% \paragraph{Why exchange helps.}
% The classifier sets \emph{what} each component should learn but, on its own,
% cannot bridge the gap between $\pi_{\theta_k}$'s current reachable set
% $\mathcal{R}^{\pi_{\theta_k}}(s_0)$ and the subregion $\{x:k^\star(x)=k\}$
% assigned to it. Exchange supplies in-subregion samples from the first iteration,
% compounding the partition's effect on coverage rather than waiting for
% $\pi_{\theta_k}$ to drift into its region. The MoGX--MoG gap should therefore
% widen on rewards whose modes are far apart in trajectory space, where un-aided
% drift is slowest, a prediction \cref{sec:experiments} confirms.

% \subsection{The Training Loop}
% \label{sec:training_loop}

% \cref{alg:mogx} (\cref{app:method-details}) gives one full MoGX iteration. Only
% three steps cross rank boundaries: the trajectory all-to-all (MoGX only), an
% optional policy-gradient all-reduce (DDP only), and the classifier-gradient
% all-reduce (MoG and MoGX). Everything else is rank-local: per-component
% parameters $\theta_k$ are never broadcast or averaged, each local buffer is
% private, and remote buffer ranks see only fire-and-forget add and score
% messages. Removing the exchange step recovers MoG; removing exchange and reward
% shaping recovers NoComm; adding a policy-gradient all-reduce recovers DDP. The
% only object coupling the community at the gradient level is a single modest
% classifier, which is what lets MoG and MoGX convert \emph{compute} into
% \emph{coverage} under a fixed query budget rather than into redundant rollouts
% of a shared stale policy.

\section{Experiments}
\label{sec:experiments}
\vspace{-2mm}

We compare MoG and MoGX with three population baselines on HyperGrid under six reward landscapes, with community sizes spanning two orders of magnitude. The rewards differ in how local or non-local their structure is, which lets us see how the landscape interacts with each method. We then turn to a genomic ideotype-design environment whose optima are placed by a real genetic map (\cref{sec:chromax}).

\paragraph{Environments.}
HyperGrid \citep{bengio2021flow} with $\texttt{ndim}{=}10$ under six rewards in two regimes. The \emph{non-tiered} rewards follow prior GFlowNet work: \emph{original} (corner modes, \citealp{bengio2021flow}, $H{=}10$, $R_0{=}0.01$), \emph{cosine} (a regular cosine-shaped pattern of high-reward regions, $H{=}16$) and \emph{sparse} (tiny isolated corner modes from GAFN, \citealp{pan2023gafn}, $H{=}16$). The \emph{tiered/compositional} rewards ($H{=}16$) make structure found in one region determine where reward lies elsewhere. In \emph{bitwise\_xor}, a global $\mathrm{GF}(2)$ parity rule, enforced at progressively stricter bit-planes, scatters modes with no spatial gradient; each policy must infer the same non-local rule, which then predicts every remaining mode. In \emph{multiplicative\_coprime}, each tier adds a different number-theoretic constraint (prime support, exponent caps, cross-coordinate coprimality, LCM targets), so policies must compose heterogeneous constraints, and factorizations learned near the origin are prerequisites for uncovering the rarer high-tier modes. In \emph{conditional\_multiscale}, each scale's constraint depends on the coarser scales already resolved (a coarse-to-fine shifted-digit filter), so modes must be found in a fixed coarse-to-fine order. The partition function grows as $H^{10}$, which puts all but the smallest grids in the mode-bias regime of \cref{sec:background}. These \emph{tiered/compositional} rewards are much harder than earlier HyperGrid variants (see the appendix of \citealp{richardson2026local} for an initial version of these environments); \cref{app:environments} gives full details.

\paragraph{Baselines.}
All five methods share the per-policy architecture, and each agent runs for $1.31\times 10^{8}$ trajectories (\cref{app:hyperparams}). They differ only in how policies coordinate. \textbf{NoComm}: $K$ independent GFlowNets. \textbf{DDP}: $K$ agents with all-reduced gradients, i.e., an effective batch of $K\cdot 1024$. \textbf{EGFN} \citep{ikram2024evolution}: rank $0$ is the only gradient-trained GFlowNet; the other $K{-}1$ ranks form a population evolved every iteration by selection, crossover and mutation, and each member sends its $64$ best trajectories to the learner's prioritized replay buffer (ported from the official code and adapted to our setup; see below and \cref{tab:hyperparams}). \textbf{MoG} (ours): classifier-driven reward shaping, with only classifier gradients all-reduced. \textbf{MoGX} (ours): MoG plus exchange of $128$ of $1024$ trajectories per step. We sweep $K\in\{2,4,8,16,32,64\}$ and read every run at the same \emph{total} budget, summed over agents: $2.62\times10^{8}$ trajectories, where a $K{=}2$ community stops. This is deliberate: we target settings where oracle queries are the scarce resource, so the question is how best to spend a fixed number of them. The cost is that agents in larger communities train for less: at $K{=}64$ each agent sees $4.1\times10^{6}$ trajectories ($4{,}000$ gradient steps), against $128{,}000$ steps at $K{=}2$. A single GFlowNet ($K{=}1$, where NoComm, DDP, MoG and MoGX coincide) stops at $1.31\times10^{8}$ trajectories, half the shared budget, so it is not in the endpoint plots. We report the cumulative number of unique modes the community finds, as mean $\pm 1$ s.d.\ over 4 seeds; discovery curves are in \cref{app:curves}.

\paragraph{Adapting EGFN.}
EGFN is hard to compare fairly. The original pairs a learner batch of $16$ with $5$ population agents sampling $4$ trajectories each. For fairness, we match the algorithm hyperparameters: $K$ networks, $1024$ trajectories per rank per iteration, and the same number of gradient steps. At this batch size EGFN has a weakness. Its mutation adds $\mathcal{N}(0,1)$ noise to whole weight tensors, which makes a population policy nearly deterministic: after one mutation, $1024$ samples cover $2$ distinct terminal states, so most of the population budget is wasted. We run EGFN only off-policy, because the population's trajectories reach the learner only through the replay buffer.

\begin{figure}[!t]
    \centering
    \includegraphics[width=\linewidth]{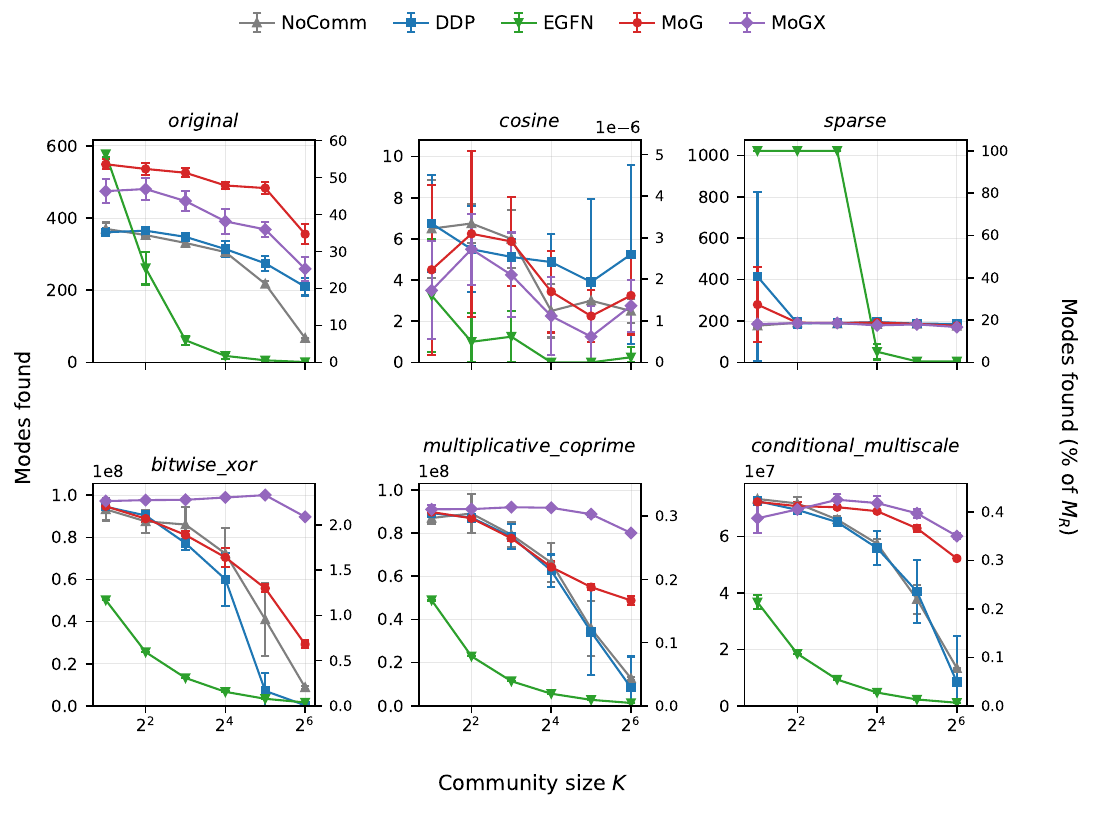}
    \caption{Modes found at a fixed total-trajectory budget vs community size $K$ (log $K$ axis; right axis: \% of all modes), \textbf{off-policy regime} (replay mix ratio $0.5$). MoG (red) and MoGX (purple) scale with $K$ while DDP (blue), NoComm (grey), and EGFN (green) tend to decline; MoGX gains most on the structured rewards.}
    \label{fig:final_modes_vs_K}
\end{figure}

\paragraph{Results (off-policy, replay mix ratio $\mathbf{\rho=0.5}$).}
Each training batch mixes fresh on-policy trajectories and replayed samples half and half. \Cref{fig:final_modes_vs_K} shows the headline result: MoG and MoGX generally improve with $K$, while NoComm and DDP use trajectories less efficiently as agents are added. The discovery curves (\cref{app:curves}) show why. NoComm plateaus quickly because independent agents rediscover the same modes, and DDP saturates soon after because every agent shares the same stale parameters and reachable set, as \citet{singla2024sapg} found for on-policy RL. MoG scales better because the soft partition pushes each component into its own subregion. MoGX generally improves over MoG, since exchange gives every component in-subregion samples from the first iteration. The gap grows with the number of modes: on \emph{conditional\_multiscale} and \emph{bitwise\_xor}, where policy bias hurts most, MoGX keep its coverage as $K$ grows while NoComm and DDP drop by up to an order of magnitude. On \emph{sparse}, no method gains from larger $K$. With few, distant modes, the bottleneck is reaching any mode at all, and partitioning cannot help with that. EGFN is strongest at small $K$: it leads on \emph{original} at $K{=}2$, and on \emph{sparse} it is the only method to find every mode, up to $K{=}8$. It then degrades quickly, because at a fixed total budget its single learner takes fewer steps as $K$ grows and the collapsed population does not make up for it. 

\begin{figure}[!t]
    \centering
    \includegraphics[width=\linewidth]{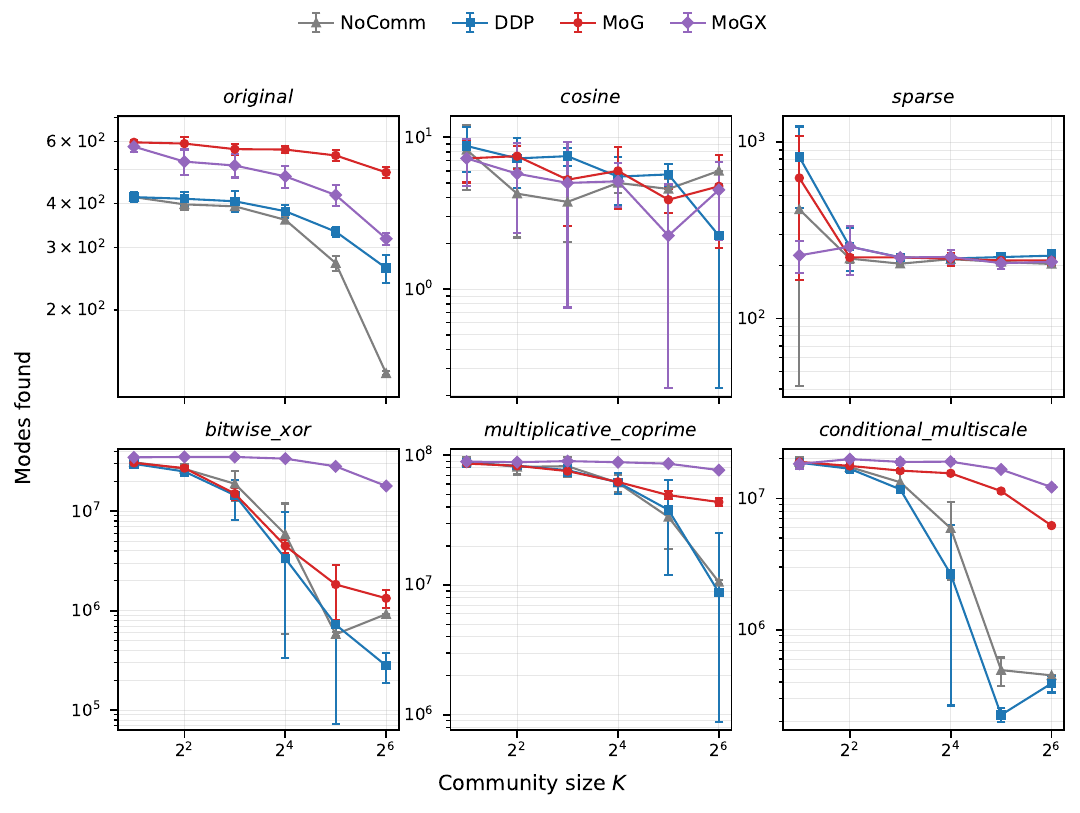}
    \caption{As \cref{fig:final_modes_vs_K}, \textbf{pure on-policy regime}. MoG and MoGX still separate from DDP and NoComm, and the MoGX--MoG gap is wider without replay.}
    \label{fig:final_modes_vs_K_online}
    \vspace{-5mm}
\end{figure}

\paragraph{Pure on-policy ($\mathbf{\rho=0.0}$).}  We rerun the sweep without a replay buffer, so every batch comes from the current policies (\cref{fig:final_modes_vs_K_online}). Dropping replay does not change how the methods scale: each method loses coverage with $K$ at almost the same rate as in \cref{fig:final_modes_vs_K}, and the methods rank in the same order. MoG and MoGX keep their lead over NoComm and DDP without replay.

\paragraph{Per-iteration cost.}
\Cref{tab:timings_total} gives the mean wall-clock time per iteration on \emph{original}, pure on-policy, over 4 seeds (per-reward and per-phase numbers in \cref{app:timings}; \cref{tab:timings_original} is the phase breakdown for the same runs). Rank-local work (sampling, the loss forward pass with reward shaping, the backward pass) stays flat in $K$ at ${\sim}700$\,ms for every method, so a larger community adds no per-rank compute. What grows with $K$ is cross-rank traffic. DDP's policy-gradient all-reduce climbs from $24$\,ms at $K{=}2$ to $549$\,ms at $K{=}64$. MoG has \emph{no} policy-gradient sync, only all-reduce on a two-hidden-layer MLP classifier. At $K{=}64$ MoG's is $22\%$ faster ($976$ vs.\ $1249$\,ms). MoGX is slower than DDP at every $K$, and at $K{=}64$ its exchange ($760$\,ms) costs more than the rank-local work. The overhead comes from our implementation, which calls mpi4py's pickle-based \texttt{alltoall}, serializing every payload as a Python object, and makes every rank wait for the slowest one. A buffer-based, asynchronous exchange would avoid both costs. Small differences between methods are run-to-run noise.

\begin{figure}[t]
    \centering
    \begin{minipage}[c]{0.64\linewidth}
        \centering
        \begin{subfigure}{0.49\linewidth}
            \includegraphics[width=\linewidth]{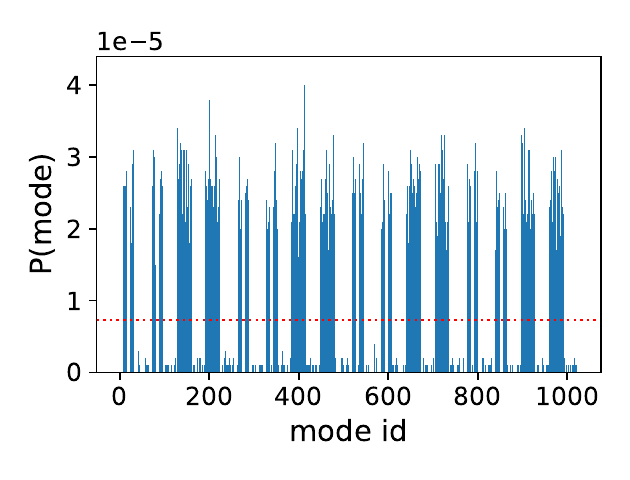}
            \caption{Rank $0$, $\log Z_0{=}12.20$,
            $\mathrm{KL}(\pi_{\theta_0}\,\|\,\pi^\star)=8.4\!\times\!10^{-3}$.}
        \end{subfigure}\hfill
        \begin{subfigure}{0.49\linewidth}
            \includegraphics[width=\linewidth]{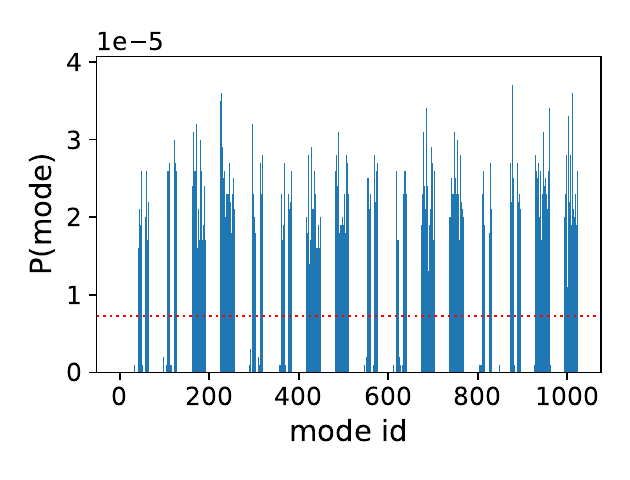}
            \caption{Rank $1$, $\log Z_1{=}12.18$,
            $\mathrm{KL}(\pi_{\theta_1}\,\|\,\pi^\star)=8.1\!\times\!10^{-3}$.}
        \end{subfigure}

        \vspace{0.5em}
        \begin{subfigure}{0.49\linewidth}
            \includegraphics[width=\linewidth]{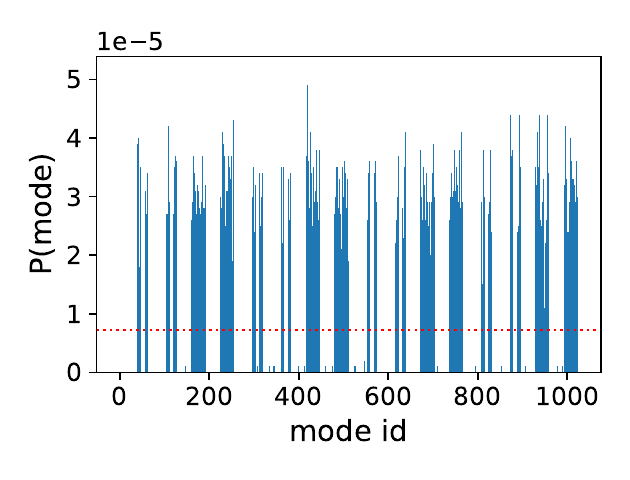}
            \caption{Rank $2$, $\log Z_2{=}11.92$,
            $\mathrm{KL}(\pi_{\theta_2}\,\|\,\pi^\star)=1.03\!\times\!10^{-2}$.}
        \end{subfigure}\hfill
        \begin{subfigure}{0.49\linewidth}
            \includegraphics[width=\linewidth]{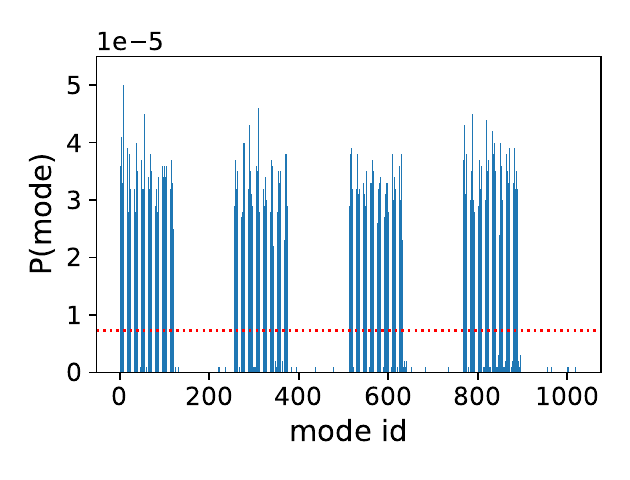}
            \caption{Rank $3$, $\log Z_3{=}11.90$,
            $\mathrm{KL}(\pi_{\theta_3}\,\|\,\pi^\star)=1.11\!\times\!10^{-2}$.}
        \end{subfigure}
    \end{minipage}\hfill
    \begin{subfigure}[c]{0.33\linewidth}
        \includegraphics[width=\linewidth]{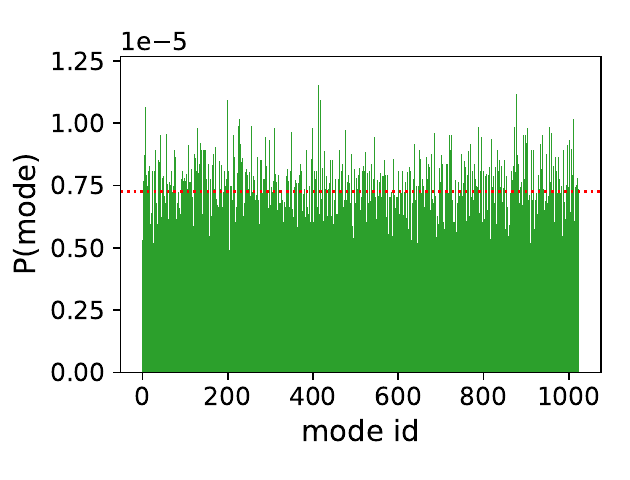}
        \caption{Z-weighted mixture,
        $\mathrm{KL}(p_\Pi\,\|\,\pi^\star)=
        1.51\!\times\!10^{-4}$.}
    \end{subfigure}
    \caption{Mixture correctness (\cref{eq:mixture-target}) for a $K{=}4$ MoGX checkpoint on HyperGrid \emph{original}. Bars are empirical per-mode marginals from $10^6$ on-policy trajectories; the dotted red line is the optimal policy $\pi^\star$ (uniform, $7.26\!\times\!10^{-6}$ per mode). Each rank covers a distinct sparse subset of modes; the Z-weighted mixture recovers $\pi^\star$.}
    \label{fig:mog_mixture_correctness}
\end{figure}

\begin{table}[h] 
\centering
\footnotesize
\setlength{\tabcolsep}{4pt}
\begin{tabular}{lrrrrrrr}
\toprule
\textbf{Method} & \textbf{$K{=}1$} & \textbf{$2$} & \textbf{$4$} & \textbf{$8$} & \textbf{$16$} & \textbf{$32$} & \textbf{$64$} \\
\midrule
\textbf{NoComm} & 702 \SE{4} & 729 \SE{6} & 726 \SE{7} & 728 \SE{7} & 732 \SE{3} & 758 \SE{23} & 713 \SE{9} \\
\textbf{DDP}    & 702 \SE{4} & 715 \SE{3} & 743 \SE{2} & 784 \SE{1} & 823 \SE{2} & 950 \SE{4} & 1249 \SE{11} \\
\textbf{EGFN}   & 702 \SE{4} & 712 \SE{3} & 748 \SE{23} & 853 \SE{14} & 876 \SE{70} & 803 \SE{6} & 975 \SE{3} \\
\textbf{MoG}    & 702 \SE{4} & 721 \SE{4} & 749 \SE{7} & 800 \SE{3} & 818 \SE{6} & 923 \SE{8} & 976 \SE{7} \\
\textbf{MoGX}   & 702 \SE{4} & 762 \SE{30} & 804 \SE{15} & 841 \SE{12} & 998 \SE{12} & 1215 \SE{49} & 1477 \SE{12} \\
\bottomrule
\end{tabular}
\caption{Mean per-iteration wall-clock time (ms) by method and $K$, pure on-policy, \emph{original} reward, over 4 seeds and all steps. NoComm stays flat, MoG grows with the non-blocking classifier all-reduce, DDP with the $\nabla_\theta$ all-reduce, and MoGX adds the exchange all-to-all. At $K{=}1$ all methods are the same loop.}
\label{tab:timings_total}
\vspace{-5mm}
\end{table}
\paragraph{Empirical validation of mixture correctness.}
The proposition in \cref{sec:overview} assumes every component exactly reaches its shaped target. We test whether a finitely trained $K{=}4$ MoGX checkpoint (\emph{original}, $\textsf{ndim}{=}10$, $H{=}7$, $1024$ corner modes) still gives $p(x)\propto R(x)$ (setup in \cref{app:mixture}). \Cref{fig:mog_mixture_correctness} shows each rank far from the optimum, as intended: each rank targets its own $\tilde R_k$ and puts mass on a distinct, sparse subset of the modes. The Z-weighted mixture nonetheless spreads mass almost uniformly over all $1024$ modes, with $\mathrm{KL}(p_\Pi\,\|\,\pi^\star)=1.51\!\times\!10^{-4}$. The learned $\log Z_k$ are accurate enough that biased, classifier-shaped sub-policies combine into an unbiased sampler.

\subsection{Genomic ideotype design}
\label{sec:chromax}
\vspace{-2mm}

Plant breeding is an AI-for-Science problem that needs a large set of high-reward states. Breeders want \emph{many} distinct genotypes that meet a target trait profile, because each candidate must be crossed, grown and phenotyped over several field seasons before its value is known \citep{tester2010breeding}, and because a genetically diverse portfolio protects against disease and climate \citep{esquinas2005protecting}. \citet{donald1968ideotypes} introduced the \emph{ideotype}, a model plant defined by the traits expected to maximize yield. Genomic ideotype design carries the idea to the DNA level: one assembles a target genotype by pyramiding beneficial alleles at known loci. Alleles close together on a chromosome are inherited together, though, and cannot be recombined freely. We evaluate MoG and MoGX in this setting.

\paragraph{Environment.}
A state is a dosage vector $x\in\{0,\dots,P{-}1\}^{L}$ over $L=B\cdot b$ loci,
starting from $\mathbf{0}_L$. Each action raises the dosage at one locus by one
(masked at the top dosage), and termination is legal at every state, so
$|\mathcal{X}|=P^{L}$. The loci are partitioned
into $B$ blocks of size $b$; a block is \emph{satisfied} when its dosage sums to
exactly $t^\star$, with any excess counting as the double-dosage failure case.
Writing $s(x)$ for the number of satisfied blocks,
\begin{equation}
\log R(x) =
\begin{cases}
\beta\,s(x) \;-\; \gamma\sum_{i=1}^{L} w_i\,x_i, & s(x)>0,\\[2pt]
\log R_0, & \text{otherwise.}
\end{cases}
\label{eq:chromax-reward}
\end{equation}
The weights $w_i=-\log\bigl(r_i+\epsilon\bigr)$ tie the landscape to biology: $r_i$ is the Haldane recombination fraction \citep{haldane1919combination} between locus $i$ and its block's anchor, read off the wheat genetic map of \citet{wang2014wheat}, as shipped with ChromaX \citep{younis2023chromax}. Tightly linked loci get large weights, so the penalty measures how improbable the needed recombination is. We use $B{=}10$, $b{=}2$, $P{=}2$, $t^\star{=}1$, $\beta{=}4$, $\gamma{=}1$.

\begin{figure}[t]
    \centering
    \begin{subfigure}{0.48\linewidth}
        \centering
        \includegraphics[width=\linewidth]{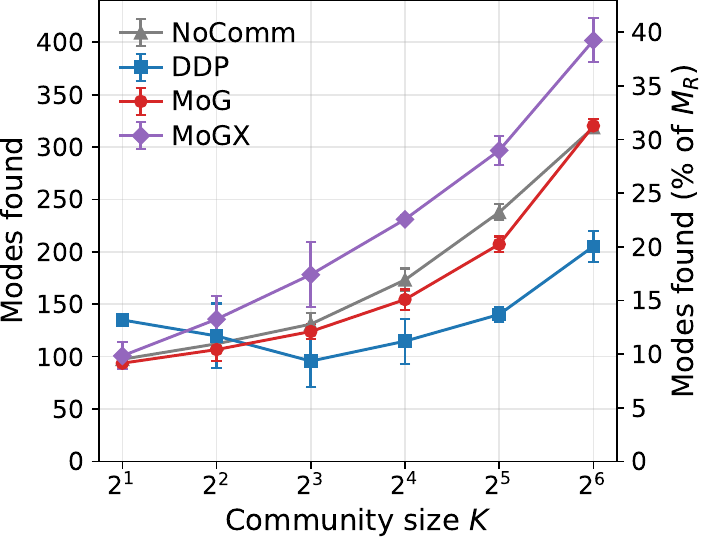}
        \caption{Replay regime ($\rho{=}0.5$).}
        \label{fig:chromax_final_modes_replay}
    \end{subfigure}\hfill
    \begin{subfigure}{0.48\linewidth}
        \centering
        \includegraphics[width=\linewidth]{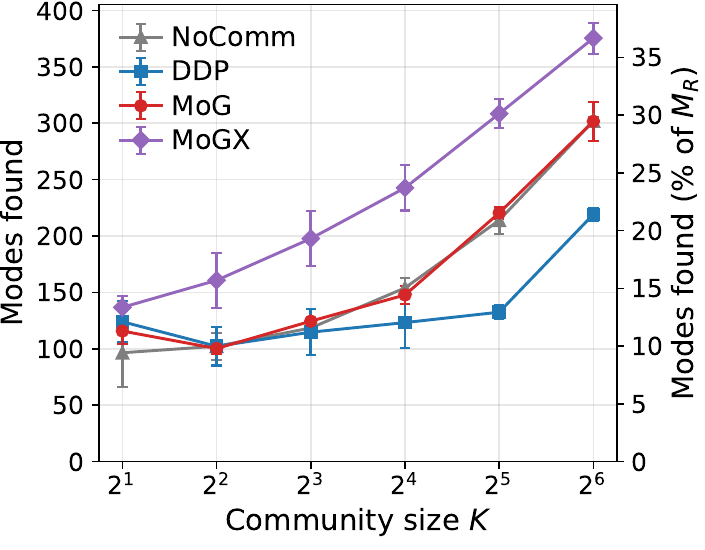}
        \caption{Pure on-policy regime ($\rho{=}0$).}
        \label{fig:chromax_final_modes_online}
    \end{subfigure}
    \caption{Modes found at a fixed total-trajectory budget vs $K$ on the ChromaX ideotype environment ($B{=}10$, $\beta{=}4$; $1024$ optima), mean $\pm 1$ s.d.\ over $4$ seeds. MoGX (purple) leads at every $K$, MoG (red) tracks NoComm (grey), and DDP (blue) is generally last. Dropping replay costs MoGX, MoG and NoComm $5{-}7\%$ of their $K{=}64$ coverage and gains DDP $7\%$; no ordering changes.}
    \label{fig:chromax_final_modes}
    \vspace{-5mm}
\end{figure}

\paragraph{Results.}
\Cref{fig:chromax_final_modes} reports modes found at a budget of $2.62\times10^{8}$ trajectories, with and without replay, over $4$ seeds (curves in \cref{fig:lc_chromax,fig:lc_chromax_online}). Both regimes agree: MoGX leads at every $K$, MoG matches NoComm, and DDP falls behind, as the mode-bias argument of \cref{sec:background} predicts.

\section{Conclusion}
\label{sec:conclusion}
\vspace{-2mm}

Larger batches and data-parallel gradient averaging do not enlarge the set of regions a distributed system can reach. \textbf{MoG} targets that reachable set directly: a community of $K$ GFlowNets, coupled through one jointly trained classifier, each learns a distinct subregion of the target. The mixture is provably correct, and we showed empirically that biased, classifier-shaped sub-policies recombine into an unbiased sampler. \textbf{MoGX} adds classifier-driven trajectory exchange, which gives each component in-subregion experience from the first iteration. With the environment-interaction budget held fixed, both methods turn parallel compute into mode coverage. Across six HyperGrid rewards and up to $64$ policies, MoG and MoGX generally keep finding modes as $K$ grows while DDP saturates and independent agents plateau, with or without replay, and MoG synchronizes more cheaply than DDP. On ideotype design, with optima separated by linkage from a real wheat map, MoGX reaches $39\%$ of the $1024$ optima at $K{=}64$, while MoG alone only matches independent agents, likely because of the reward's geometry.

Two limitations point to future work. Both environments share the same state-space structure, so we have not tested whether a learned partition works for graph or sequence generation with larger action spaces and variable trajectory lengths. And the community is static: $K$ is fixed in advance, with no way to retire or reallocate stalled components; a discovery-rate signal could drive that resizing. The obvious next step is molecule and biological-sequence design, where oracle calls are expensive and structural diversity is the point. The framework does not depend on the balance objective or the form of the partition, and we expect its core idea, giving each parallel agent its own learned sub-target, to carry over.

\newpage

% \section{Ethics Statement}
% \label{sec:ethics}

% This work uses public protein sequence and structure data and does not involve human subjects, private personal data, or clinical deployment. The remaining risk is dual-use: generated binders should not be synthesized, tested, or deployed without appropriate biosafety review, target-risk assessment, experimental validation, and regulatory oversight. Our models inherit biases from SAbDab, UniRef100, and the learned reward models, including biases toward well-studied antibody families, antigens, and assay conditions. This work should not be considered safe to use in any autonomous discovery program.

\section{Reproducibility Statement}
\label{sec:reproducibility}

We will release all code, including model training code, environments, and analysis scripts. These artifacts are intended to allow the community to inspect the work, extend the framework to new domains, and benchmark the method easily on more challenging problems.

\section{AI Assistance Disclosure}
\label{sec:ai-assistance}

Large language models (LLMs) were used to assist with manuscript drafting (primarily consistency checks, accuracy of claims validation, and LaTeX automation), and for code generation for  model evaluations. Many elements of this codebase predate the advent of human-level coding agents, so most core methods were implemented entirely by hand. LLMs were used under heavy supervision for figure and table generation. They were not used for research ideation, method design, or experiment design. The authors reviewed all code underlying reported results and rewrote any generated manuscript text by hand.

\newpage

\bibliography{iclr2027_conference}
\bibliographystyle{iclr2027_conference}

\newpage
\appendix

\section{Extended Related Work}
\label{app:related}

\paragraph{GFlowNets, MaxEnt RL and continuous samplers.}
The mode collapse discussed in \cref{sec:background} is specific to return
maximization. \emph{Entropy-regularized} (MaxEnt) RL instead learns a stochastic
policy that samples in proportion to reward, and GFlowNet training can be recast
as a MaxEnt RL problem with a particular reward and regularizer
\citep{tiapkin2024generative}. In the multi-path environments typical of
sequential construction, this equivalence requires correcting the reward so that
the optimal policy's marginal is proportional to $R(x)$ regardless of the
underlying MDP \citep{deleu2024discrete}. Sampling from an unnormalized density
is also the aim of diffusion and stochastic-control samplers
\citep{zhang2022path,vargas2023denoising}, to which continuous GFlowNets
\citep{lahlou2023continuous} are closely related. We use GFlowNets because they
match the multi-path, discrete structure of our construction processes, not
because we see them as a separate paradigm.

\paragraph{Parallel and population-based training in RL.}
Distributed actor-learner architectures such as A3C \citep{mnih2016a3c}, IMPALA
\citep{espeholt2018impala}, and Ape-X \citep{horgan2018distributed} scale
experience collection by spreading actors across many workers. Only
IMPALA corrects for the resulting lag between actor and learner policies, with
V-trace importance weights; A3C and Ape-X apply no off-policy correction. These
frameworks parallelize a single policy and do not enforce specialization across
subregions of the target. Population-based training
\citep{jaderberg2017population} maintains a population of agents with diverse
hyperparameters, periodically replacing under-performers; diversity-aware
variants additionally regularize for behavioral spread
\citep{parkerholder2020dvd}. Specialization there emerges only implicitly from
hyperparameter perturbations or behavioral diversity bonuses, not from an
explicit partition of the target distribution, and connects to a broader
literature linking GFlowNet training to maximum-entropy reinforcement learning
\citep{tiapkin2024gfn_entropy,deleu2024discrete}. Mixture-of-experts
architectures \citep{jacobs1991adaptive,shazeer2017outrageously,ceron2024moe_rl}
are closer in spirit: a learned gate routes inputs to specialized sub-networks.
Our classifier-driven partition can be viewed as a mixture of experts in which
the experts are full generative policies rather than feed-forward heads, and the
gate's role is to define each expert's \emph{target distribution} rather than to
combine their outputs.

\paragraph{Parallel and distributed GFlowNets.}
Several recent methods parallelize GFlowNet training. Only one of them also
targets mode coverage. Trajectory Balance with Asynchrony
\citep{bartoldson2025trajectory} decouples exploration from learning by
spreading off-policy searchers across nodes that populate a shared replay
buffer, yielding large speedups for language-model post-training; the searchers,
however, are snapshots of a single policy, so the underlying reachable set is
shared and the population is asymptotically homogeneous. Embarrassingly Parallel
GFlowNets \citep{da2024embarrassingly} and federated contrastive variants
\citep{dasilva2024federated} train one local GFlowNet per factor of the
\emph{reward} (or per data shard in federated Bayes) and aggregate them via
balance conditions that recover a global target, so that reward evaluation can
be spread across machines. Subgraph Asynchronous Learning (SAL,
\citealp{sal2025distributed}) comes closest to our setting. It splits the
\emph{state graph} into subgraphs, trains one GFlowNet per subgraph in parallel,
and then combines them. The authors report up to $8\times$ more modes than a
single GFlowNet. In SAL the split is chosen before training and the
local GFlowNets train in isolation; in MoG the partition is learned jointly with
the policies, and in MoGX they also exchange trajectories. Compositional Sculpting \citep{garipov2023compositional} composes
pre-trained generative processes via classifier-guided routing, but operates
post-hoc on fixed models rather than co-training a community of policies. \emph{No existing method scales GFlowNet training by partitioning
the state space across a community of policies that are trained jointly under a
fixed environment-interaction budget}, precisely the regime that matters for AI
for Science, where the dominant per-sample cost is an expensive environment
query rather than gradient compute.

\section{Shared Distributed Infrastructure}
\label{app:infra}

Every method in this paper (NoComm, DDP, EGFN, MoG, MoGX) runs on the same distributed
stack. Isolating this stack as a single shared layer serves two purposes: it lets
us hold the environment-interaction budget fixed across methods and community
sizes, so that performance differences reflect the method rather than uneven
sampling; and it provides a single place to define the global mode registry that
produces our headline metric.

\paragraph{Rank topology.}
A run with community size $K$ uses $K$ \emph{training ranks}, $B$ \emph{buffer
ranks} (default $B{=}1$), and an optional \emph{coordinator rank}. Training ranks
own a GFlowNet replica $\theta_k$ plus a thin local replay buffer of capacity
$2048$; buffer ranks own a remote replay buffer plus per-rank
mode-discovery state; the coordinator, when present, maintains the union mode set
across the population. Training and coordinator ranks share an MPI
sub-communicator used for all collectives (all-reduce of $f_\psi$ gradients,
all-to-all of exchanged trajectories). Buffer ranks are excluded from the
sub-communicator so they never participate in training-time synchronization.

\paragraph{Replay and mode registry.}
Each training rank composes the loss-time batch as a fixed mixture of fresh
on-policy trajectories and replay samples. After loss evaluation,
the batch is shipped to the assigned buffer rank for centralized scoring; the
buffer rank invokes the environment's mode-detection routine on every
terminating state, hashes any newly discovered modes, and updates a per-rank
counter $n_k^{(t)}$. The coordinator unions these counters and exposes the
global metric \texttt{n\_modes\_found} reported in \cref{sec:experiments}. For
MoG and MoGX, the mode registry is read-only; the same infrastructure supports
discovery-rate-driven replacement variants of the
framework that are out of scope for this paper. \Cref{app:objectives} repeats the on-policy sweep with SubTB and DB.

\section{Method Implementation Details}
\label{app:method-details}

\paragraph{Loss-agnostic instantiation.}
The framework is orthogonal to the choice of GFlowNet objective. We instantiate
with trajectory balance (TB) \citep{malkin2022trajectory} throughout, but the
same code path applies to detailed balance, sub-trajectory balance
\citep{madan2023learning}, log-partition variance, and modified DB. The only
requirement is that the loss be off-policy-friendly when consumed through a
replay buffer, a property shared by all four objectives since each is computed
from $(s,a,s')$ tuples or trajectories without an importance-weighted likelihood
correction tied to the sampling policy.

\paragraph{Classifier coherence.}
$f_\psi$ consumes the same $K$-hot preprocessed state
$\phi(x)\in\mathbb{R}^{\textsf{ndim}\cdot\textsf{height}}$ used by the policies.
Its parameters are broadcast from rank $0$ at construction time, so every replica
starts from identical weights; combined with the gradient all-reduce, this
gives true data-parallel training of $f_\psi$ regardless of how the policies are
synchronized. Because every rank starts from the same $\psi^{(0)}$ and applies
the same averaged gradient, all replicas of $f_\psi$ remain identical
throughout training, with no broadcast beyond initialization. This is essential:
the shaped reward in \cref{eq:shaped-reward} would be incoherent across ranks if
their classifiers drifted apart.

\paragraph{Reward-shaping safety net.}
We clip the shaping term in \cref{eq:shaped-reward} to
$\mathrm{clip}\bigl(\log f_\psi(x)_k,\,-20,\,0\bigr)$, a numerical safety net:
deep into training, $\log f_\psi(x)_k$ for an out-of-subregion state can otherwise
saturate the loss for that sample. Shaping is applied \emph{after} the
replay-buffer sampler, so the multiplier $f_\psi(x)_k$ always reflects the
current classifier weights even when $x$ was buffered many iterations ago, which
is what makes joint training stable, and it is computed inside a
\texttt{torch.no\_grad()} block so policy gradients never propagate back into the
classifier.

\paragraph{Self-consistency.}
The all-reduced cross-entropy in \cref{eq:fpsi-loss} is equivalent to supervised
classification on the union dataset
$\bigcup_k \{(x,k):x\in\mathcal{B}^{(k)}\}$. The Bayes-optimal classifier under
that distribution is
\[
f_\psi^\star(x)_k
\;=\;
\Pr(\text{component}{=}k\mid x)
\;=\;
\frac{\pi_{\theta_k}(x)\,\mathcal{P}_\Pi(\pi_k)}
     {\sum_j \pi_{\theta_j}(x)\,\mathcal{P}_\Pi(\pi_j)}.
\]
As each $\pi_{\theta_k}$ specializes onto a subregion of $\mathcal{X}$, the
classifier sharpens its routing of states to whichever component covers them
most, and \cref{eq:shaped-reward} re-amplifies that specialization on the next
iteration. The two halves close the loop, and \cref{eq:mixture-target} is the
fixed point.

\paragraph{Symmetry breaking and initialization.}
At iteration $0$, $f_\psi$ outputs a near-uniform softmax, so
$\log f_\psi(x)_k\approx -\log K$ pointwise and \cref{eq:shaped-reward} reduces to
a global constant offset on every component's log-reward; this constant is
absorbed into $\log Z_k$ within a handful of gradient steps and the policy
distribution $\pi_{\theta_k}(x)\propto\tilde R_k(x)$ is unaffected. Which
subregion ends up assigned to which component is supplied by the independence of
on-policy sampling early in training. Empirically the partition stabilizes within
the first few thousand iterations and stays put.

\paragraph{Exchange transport.}
In MoGX, selected trajectories are grouped by destination rank, serialized into a
CPU-tensor payload (states, actions, terminating indices, log-rewards,
log-probabilities, estimator outputs), and exchanged via a single
\texttt{MPI\_Alltoall} on the training sub-communicator. Every rank participates
on every iteration; empty send slots are no-ops. Received payloads are concatenated with the locally retained slice of the source
batch, and enter the replay buffer as ordinary samples. The exchange step sits
between on-policy sampling and the optimization step, so the receiving
rank attributes any mode discoveries to itself and the reward shaping of
\cref{eq:shaped-reward} applies uniformly to local and received trajectories.

\section{Implementation Details and Hyperparameters}
\label{app:hyperparams}

This appendix consolidates the architecture, optimizer, and
training-schedule settings shared across all four population
configurations (NoComm, DDP, MoG, MoGX). All methods use an identical
per-agent architecture and run each agent for the same number of
trajectories. \Cref{sec:experiments} compares them at a shared
\emph{total} budget, so agents in larger communities take fewer
gradient steps. The methods differ only in their inter-agent coordination,
as described in \cref{sec:methods}. \cref{tab:hyperparams} lists the full
configuration. All experiments are implemented on top of the
\texttt{torchgfn} library \citep{viviano2023torchgfn}.

\paragraph{Hardware.}
All runs are executed on the CPU partition of the Trillium cluster
(SciNet, University of Toronto; part of the Digital Research Alliance of
Canada). Each CPU node comprises two 96-core AMD EPYC ``Zen5'' (Turin)
processors for $192$ cores and $768$\,GB of RAM, connected by an NDR
InfiniBand (Mellanox Dragonfly+) fabric. We map one MPI rank to one
physical CPU core, so a community of size $K$ uses $K$ cores for the
training ranks plus one core for the buffer rank
(\cref{app:method-details}). Every configuration in the sweep, up to the
largest ($K{=}64$, i.e.\ $65$ ranks), therefore fits within a single
node, and all cross-rank communication reported in \cref{app:timings}
(gradient/classifier all-reduce and the MoGX trajectory all-to-all) is
intra-node shared-memory MPI rather than inter-node network traffic.

\begin{table}
\centering
\footnotesize
\setlength{\tabcolsep}{8pt}
\begin{tabular}{ll}
\toprule
\textbf{Component / hyperparameter} & \textbf{Value} \\
\midrule
\multicolumn{2}{l}{\emph{Policy network}} \\
\quad Architecture                 & MLP, 2 hidden layers, width 256 \\
\quad Forward/backward policy      & shared parameters (tied $P_F$/$P_B$) \\
\quad GFlowNet loss                & trajectory balance (TB) \\
\quad Sampling temperature / $\epsilon$ & $T{=}1$, $\epsilon{=}0$ (on-policy) \\
\midrule
\multicolumn{2}{l}{\emph{Optimization}} \\
\quad Optimizer                    & AdamW \\
\quad Learning rate                & $10^{-3}$ \\
\quad Batch size                   & $1024$ trajectories \\
\quad LR warmup                    & linear over first $500$ iterations \\
\quad Gradient clipping            & $\|\nabla_\theta\mathcal{L}\|_2 \leq 10$ \\
\quad Per-agent budget             & $1.31\times 10^{8}$ trajectories \\
\midrule
\multicolumn{2}{l}{\emph{Partition classifier $f_\psi$ (MoG, MoGX)}} \\
\quad Architecture                 & MLP, 2 hidden layers, $K$ logits \\
\quad Input                        & $K$-hot preprocessed state $\phi(x)$ \\
\quad Initialization               & broadcast from rank $0$ (identical) \\
\quad Gradient synchronization     & all-reduce (sum-then-divide), AdamW step \\
\quad Entropy weight $\beta$        & $0$ (default) \\
\quad Shaping clip                 & $\log f_\psi(x)_k$ clipped to $[-20, 0]$ \\
\midrule
\multicolumn{2}{l}{\emph{Replay and infrastructure}} \\
\quad Local replay buffer capacity & $2048$ \\
\quad Remote replay buffer capacity & $8192$ \\
\quad Buffer ranks $B$             & $1$ (default) \\
\quad Replay mix ratio $\rho$       & $0.5$ (replay) / $0$ (pure on-policy) \\
\midrule
\multicolumn{2}{l}{\emph{Trajectory exchange (MoGX only)}} \\
\quad Exchange count $N$           & $128$ of $1024$ ($12.5\%$) \\
\quad Routing rule                 & $\arg\max_j \log f_\psi(x)_j$ \\
\midrule
\multicolumn{2}{l}{\emph{EGFN only (our adaptation, see \cref{sec:experiments})}} \\
\quad Learner / population         & rank $0$ / ranks $1,\dots,K{-}1$ \\
\quad Generations                  & one per gradient step \\
\quad Elite fraction / tournament size & $0.2$ / $4$ \\
\quad Crossover probability        & $0.1$ \\
\quad Mutation                     & $\mathcal{N}(0,1)$ per tensor, prob.\ $0.9$ \\
\quad Prioritized replay buffer    & top $1000$ by reward \\
\quad Replay sampling              & $50\%$ from top $10\%$, $50\%$ from rest \\
\quad Trajectories sent per member & top $64$ of $1024$ \\
\quad Learner batch                & $1024$ on-policy $+$ $1024$ replay \\
\midrule
\multicolumn{2}{l}{\emph{Sweep}} \\
\quad Environment                  & HyperGrid, $\textsf{ndim}{=}10$ \\
\quad Community sizes $K$           & $\{1,2,4,8,16,32,64\}$ \\
\bottomrule
\end{tabular}
\caption{Shared per-agent hyperparameters across all four population
configurations. The four methods differ only in inter-agent
coordination (\cref{sec:methods}); all other settings are identical,
which is what makes the iso-budget comparisons in
\cref{sec:experiments} meaningful. Reward-specific grid heights $H$ are
given in \cref{sec:experiments}. The EGFN rows apply to that baseline only.}
\label{tab:hyperparams}
\end{table}

\subsection{HyperGrid environments}
\label{app:environments}

All environments instantiate the $d$-dimensional HyperGrid of
\citet{bengio2021flow}: a grid of height $H$ whose states are vectors
$s\in\{0,\dots,H-1\}^{d}$, with all trajectories starting at $\mathbf{0}_d$ and
permitted to terminate in any state, giving $|\mathcal{X}|=H^{d}$ terminal
states. Where convenient we write the per-dimension normalized coordinate
$a_i(s)=\bigl|\,s_i/(H-1)-\tfrac12\,\bigr|\in[0,\tfrac12]$. Unless noted, we use
$d=10$. The three compositional rewards share a \emph{cumulative tiered} form
\begin{equation}
R(s)=R_0+\sum_{t} w_t\,\mathbf{1}\!\left[\text{$s$ satisfies all constraints of tiers $0,\dots,t$}\right],
\label{eq:tiered}
\end{equation}
with strictly increasing weights $w_t$, so that a state attains the reward of
tier $t$ only if it also satisfies every lower tier. Because the tier indicators
are cumulative, summing \eqref{eq:tiered} over $\mathcal{X}$ gives a closed-form
partition function
\begin{equation}
Z \;=\; \sum_{s} R(s) \;=\; R_0\,H^{d} \;+\; \sum_{t} w_t\,m_t,
\qquad m_t := \bigl|\{s : \text{$s$ passes tiers $0..t$}\}\bigr|,
\label{eq:tiered-Z}
\end{equation}
where $m_t$ is the tier-$t$ mode count given in closed form for each environment
below. For $H=16,\,d=10$ we have $H^{d}\approx1.10\times10^{12}$, and the
$R_0H^{d}$ term dominates $Z$ (e.g.\ $R_0{=}0.1$ gives $R_0H^{d}\approx
1.10\times10^{11}$); since the highest tier occupies a vanishing fraction of
$\mathcal{X}$, on-policy sampling almost never lands on a top-tier mode, which is
precisely the mode-bias regime of \cref{sec:background}.

\paragraph{Original \citep{bengio2021flow}.}
\begin{equation}
R(s)=R_0+R_1\prod_{i=1}^{d}\mathbf{1}[a_i(s)>0.25]
      +R_2\prod_{i=1}^{d}\mathbf{1}[0.3<a_i(s)<0.4].
\end{equation}
We use $R_0{=}0.01,\,R_1{=}0.5,\,R_2{=}2.0$ at $H{=}10$. Modes are the states
lying in the inner band $\{0.3<a_i<0.4\}$ in \emph{every} dimension
simultaneously (reward $R_0{+}R_1{+}R_2$); with $b$ admissible band indices per
axis the mode set has size $b^{d}$ (e.g.\ $b{=}2$ at $H{=}10$, i.e.\ $2^{10}$
modes concentrated near the $2^{d}$ corners).

\paragraph{Cosine.}
With $f(a)=\bigl(\cos(50a)+1\bigr)\cdot\tfrac{1}{\sqrt{2\pi}}e^{-\frac12(5a)^2}$,
\begin{equation}
R(s)=R_0+R_1\prod_{i=1}^{d} f\!\bigl(a_i(s)\bigr),
\end{equation}
with $R_0{=}0.1,\,R_1{=}0.5$ at $H{=}16$. The oscillatory factor produces a
regular lattice of high-reward regions; a state is counted as a mode when the
product exceeds a $\gamma$-scaled per-dimension peak,
$R \ge R_0 + (\gamma\, f_{\max})^{d} R_1$ (we use $\gamma{=}0.8$).

\paragraph{Sparse \citep{pan2023gafn}.}
Let the target set $\mathcal{T}$ contain, for each $k\in\{0,\dots,d-1\}$, every
distinct permutation of the vector with $k$ entries equal to $1$ and $d-k$
entries equal to $H-2$. The reward places an isolated spike on each target,
\begin{equation}
R(s)=\varepsilon+\sum_{t\in\mathcal{T}}\mathbf{1}[s=t],
\qquad \varepsilon>0 .
\end{equation}
There are $|\mathcal{T}|=\sum_{k=0}^{d-1}\binom{d}{k}=2^{d}-1$ such spikes
($1023$ at $d{=}10$; the all-ones vector is excluded), each surrounded by reward $\varepsilon$, making the modes
extremely small and maximally isolated.

\paragraph{Bitwise XOR (compositional; rule reuse).}
Let $\mathrm{bit}_b(\cdot)$ denote the $b$-th bit and restrict attention to a
subset $D\subseteq\{1,\dots,d\}$ of $|D|=m$ constrained dimensions. For tier $t$
with bit window $[\ell_t,h_t]$, the constraint requires, at \emph{every}
bit-plane $b\in[\ell_t,h_t]$, a system of $\mathrm{GF}(2)$ (XOR) parity checks
$A_t\,u^{(b)}\equiv c_t \pmod 2$ on the bit vector
$u^{(b)}\in\{0,1\}^{m}$ across constrained dimensions (defaulting to even parity
$\sum_{i\in D}\mathrm{bit}_b(s_i)\equiv 0$ when unspecified). The reward is
\eqref{eq:tiered} with these cumulative bit-plane constraints. Modes are
therefore \emph{non-local}: they are placed by an algebraic law rather than by
proximity, and carry no spatial gradient. For a single even-parity check across
$m$ dimensions at all bit-planes, one coordinate is fully determined by the XOR
of the others, and (since $H$ is a power of two the determined value is always
in range) the mode count is $H^{m-1}\cdot H^{d-m}=H^{d-1}$; e.g.\
$m{=}d{=}10,\,H{=}16$ gives $16^{9}\approx6.87\times10^{10}$ modes at the
strictest tier. The intended challenge is \emph{rule reuse}: the same parity
computation, learned once, is reused at each tier under increasing strictness, so
that identifying the rule from a handful of modes predicts the location of all
the others.

\paragraph{Multiplicative Coprime (compositional; knowledge composition).}
Coordinates are internally shifted by $+1$ (state value $0\!\mapsto\!1$) so the
origin is valid and short trajectories immediately encounter small
prime-factorable numbers, providing early signal. Over a small prime set
$P$ and active dimensions $A\subseteq\{1,\dots,d\}$, successive tiers impose
\emph{qualitatively different} constraints:
\begin{itemize}
  \item \textbf{tier $0$ (prime support):} each $s_i,\,i\in A$, factors entirely over $P$;
  \item \textbf{tier $\ge 1$ (exponent caps):} every prime exponent is bounded, $e_{i,p}\le c_t$, with $c_t$ tightening per tier;
  \item \textbf{coprimality:} designated pairs $(i,j)$ become coprime, $\gcd(s_i,s_j)=1$;
  \item \textbf{LCM targets:} an optional global $\mathrm{lcm}_{i\in A} s_i$ target.
\end{itemize}
The reward is \eqref{eq:tiered} with these cumulative constraints. Ignoring the
cross-dimensional coprimality/LCM couplings, the prime-support-plus-caps mode
count factorizes as $N^{|A|}\cdot H^{d-|A|}$, where $N$ is the number of
admissible ``smooth'' values in $\{1,\dots,H\}$; for $P=\{2,3,5\},\,c=2,\,H=16$
we have $N=10$ (namely $\{1,2,3,4,5,6,9,10,12,15\}$), giving $10^{10}$ modes when
all dimensions are active. Coprimality couples dimensions and further reduces
this count (it no longer factorizes). The intended challenge is \emph{knowledge
composition}: each tier is a different kind of number-theoretic reasoning, and
the factorization structure learnable near the origin is a prerequisite for the
coprimality and LCM reasoning that gates the rarer high-tier modes.

\paragraph{Conditional Multi-Scale (compositional; conditional hierarchy).}
Decompose each coordinate in base $B$ into $L=\log_B H$ digits,
$s_i=\sum_{k=0}^{L-1} d_k(i)\,B^{k}$. Ordering tiers coarse-to-fine, tier $t$
constrains digit $L{-}1{-}t$ (the $(t{+}1)$-th most significant) through a
\emph{shifted} filter whose offset depends on all coarser digits already
constrained:
\begin{equation}
\bigl(d_{L-1-t}(i)+\sigma_t(i)\bigr)\bmod B < f,
\qquad
\sigma_t(i)=\sum_{k=0}^{t-1} a_{t,k}\,d_{L-1-k}(i)\ \bmod B,
\end{equation}
with seed-derived coefficients $a_{t,k}$ (tier $0$ has $\sigma_0{=}0$), and an
optional cross-dimensional term $\sum_{i} d_{L-1-t}(i)\equiv 0 \pmod{m_t}$. The
reward is \eqref{eq:tiered}. Because coarse-scale digits are small near the
origin (high digits equal $0$ and pass trivially), difficulty is
distance-correlated. The mode count admits a closed form,
\begin{equation}
m_T=\bigl(f^{T}\bigr)^{d}\, B^{(L-T)d}
     \Big/ \textstyle\prod_{t} m_t ,
\end{equation}
(with the $\prod_t m_t$ factor present only under cross-dimensional
constraints), and $Z$ follows from \eqref{eq:tiered-Z} with no enumeration. As a
reference instantiation, $B{=}2,\,L{=}4,\,f{=}1,\,d{=}10$ yields
$m_1\approx1.07\times10^{9}$, $m_2\approx1.05\times10^{6}$, and $m_3{=}1024$ as
tiers accumulate. The intended challenge is a \emph{conditional hierarchy}: the
constraint at each scale is a function of the scales below it, so exploration
must proceed coarse-to-fine, with the lower tiers acting as foothills that
bootstrap discovery of the higher ones.

\section{Mixture-Correctness Validation: Setup and Full Results}
\label{app:mixture}

\paragraph{Setup.}
We take a $K{=}4$ MoGX checkpoint on HyperGrid \emph{original}
($\textsf{ndim}{=}10$, $H{=}7$, $R_0{=}10^{-8}$, ``challenging'' preset;
$2^{10}{=}1024$ corner modes; analytical reward-proportional optimum uniform
across modes at $7.26\!\times\!10^{-6}$ each with $99.26\%$ off-mode mass). For
each rank $k$ we draw $10^6$ on-policy trajectories and
bin terminating states by mode index. We then build the Z-weighted mixture
marginal $p_\Pi(x)=\sum_k w_k\,\pi_{\theta_k}(x)$ with
$w_k=\mathrm{softmax}(\log Z_k)_k$, reading $\log Z_k$ directly from the trained
TB scalar exactly as prescribed by \cref{eq:mixture-target}. The reference
optimum is computed in closed form by enumerating only the $2^{10}$ mode-band
states and using \emph{original}'s outer-ring decomposition for the partition
function $Z$ (no $H^{\textsf{ndim}}$ enumeration). KL divergences are reported
over the full $n_{\text{modes}}$-category support.

\paragraph{Full results.}
\Cref{fig:mog_mixture_correctness} reports the per-rank and mixture marginals.
Each individual rank is far from the optimum on its own, by design: each
$\pi_{\theta_k}$ targets its shaped reward $\tilde R_k=R\cdot f_\psi(\cdot)_k$,
not $R$. All four ranks place mass on distinct, sparse subsets of the $1024$
modes (off-mode mass $99.1$--$99.6\%$, forward KL
$0.81$--$1.11\!\times\!10^{-2}$). The Z-weighted mixture, in contrast, recovers
the analytical optimum: $\mathrm{KL}(p_\Pi\,\|\,\pi^\star)=1.51\!\times\!10^{-4}$,
lower than every individual rank by a factor of $50$--$75\times$, with off-mode
mass $99.31\%$ matching the analytic $99.26\%$ to within sampling noise and
on-mode mass distributed near-uniformly across the $1024$ modes. No single rank
covers the modes on its own; the mixture does, because the union of the four
ranks' coverage, weighted by $\{Z_k\}$, reassembles into the full target. This
is the operational content of MoG's mixture-correctness proposition.

\section{Full Training Curves}
\label{app:curves}

\begin{figure}[h]
    \centering
    \includegraphics[width=\linewidth]{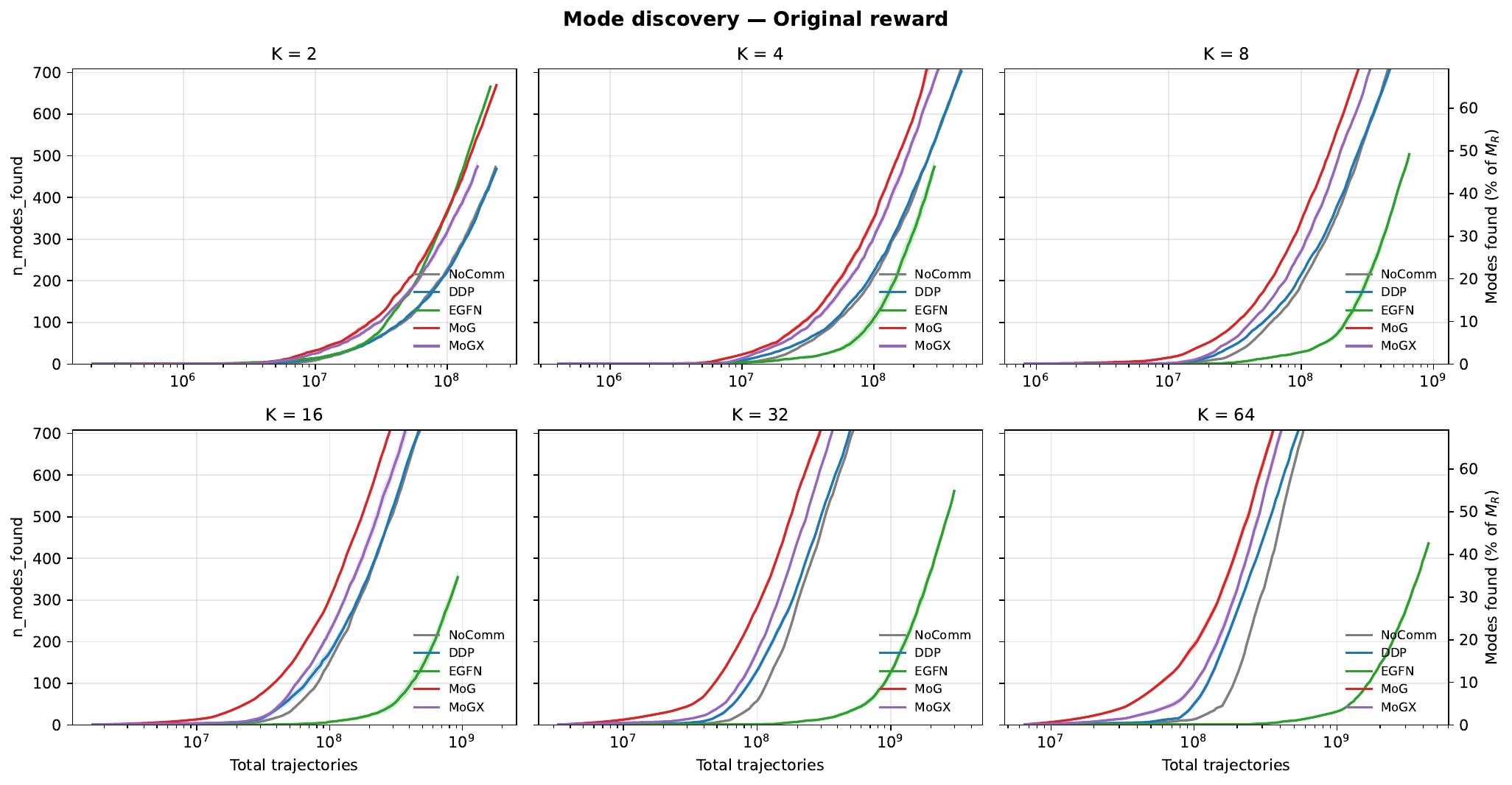}
    \caption{Modes discovered vs total trajectories on the
    \emph{original} reward, \textbf{replay regime} ($\rho{=}0.5$).
    MoGX (purple) and MoG (red) discover modes faster than DDP (blue) and NoComm (grey).}
    \label{fig:lc_original}
\end{figure}

\begin{figure}[h]
    \centering
    \includegraphics[width=\linewidth]{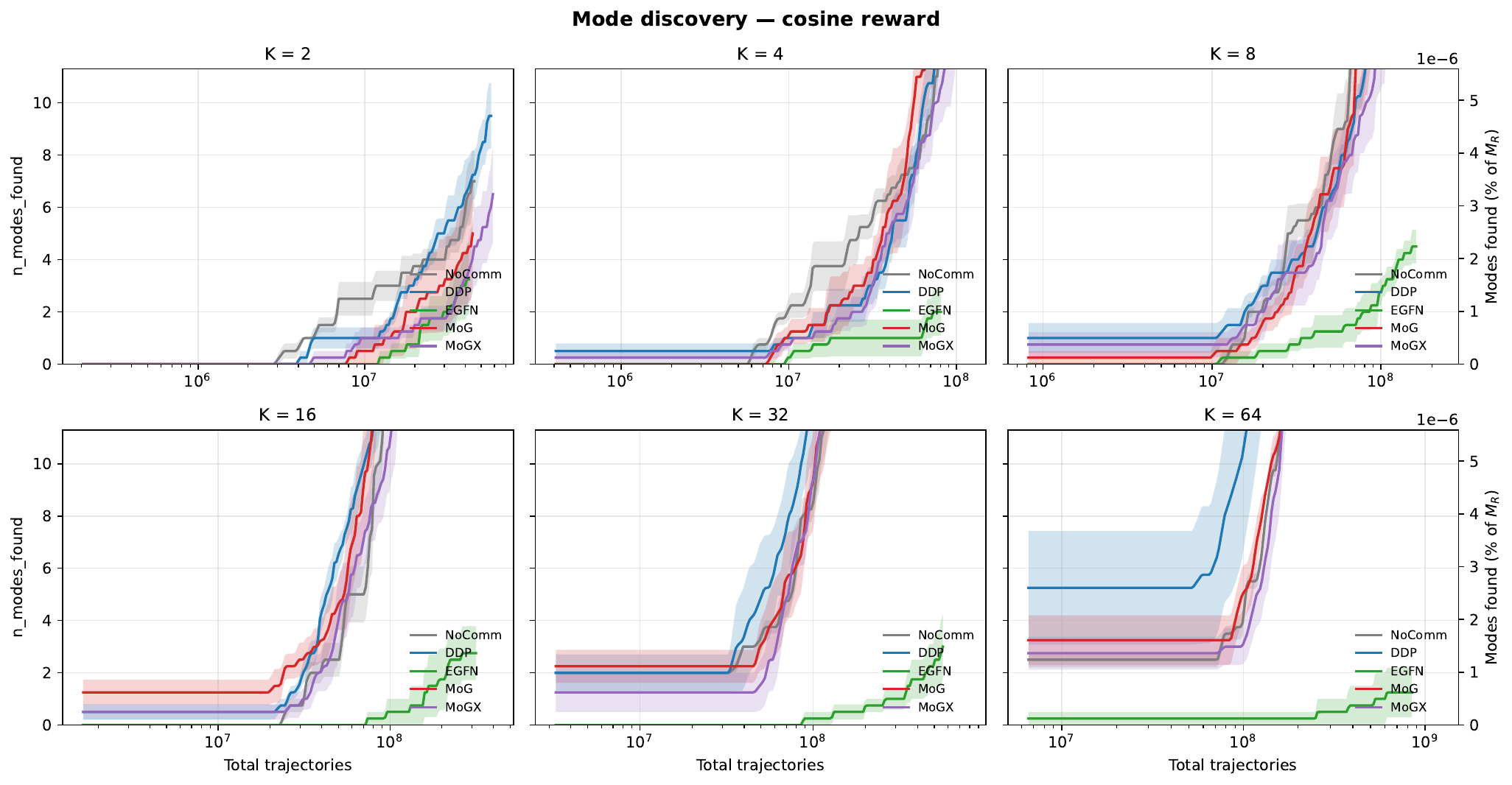}
    \caption{Modes discovered vs total trajectories on the \emph{cosine}
    reward, \textbf{replay regime} ($\rho{=}0.5$).}
    \label{fig:lc_cosine}
\end{figure}

\begin{figure}[h]
    \centering
    \includegraphics[width=\linewidth]{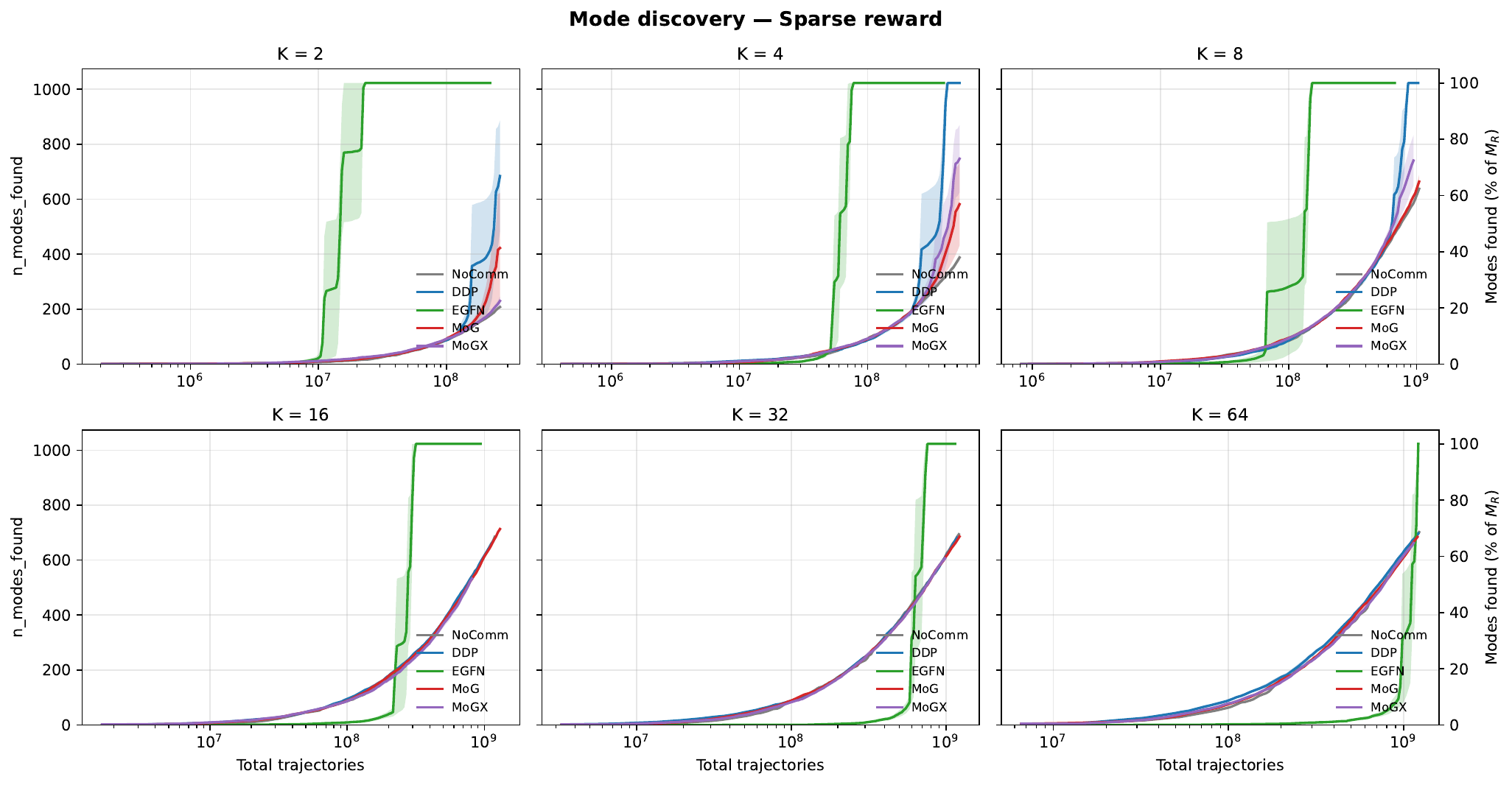}
    \caption{Modes discovered vs total trajectories on the \emph{sparse}
    reward, \textbf{replay regime} ($\rho{=}0.5$).}
    \label{fig:lc_sparse}
\end{figure}

\begin{figure}[h]
    \centering
    \includegraphics[width=\linewidth]{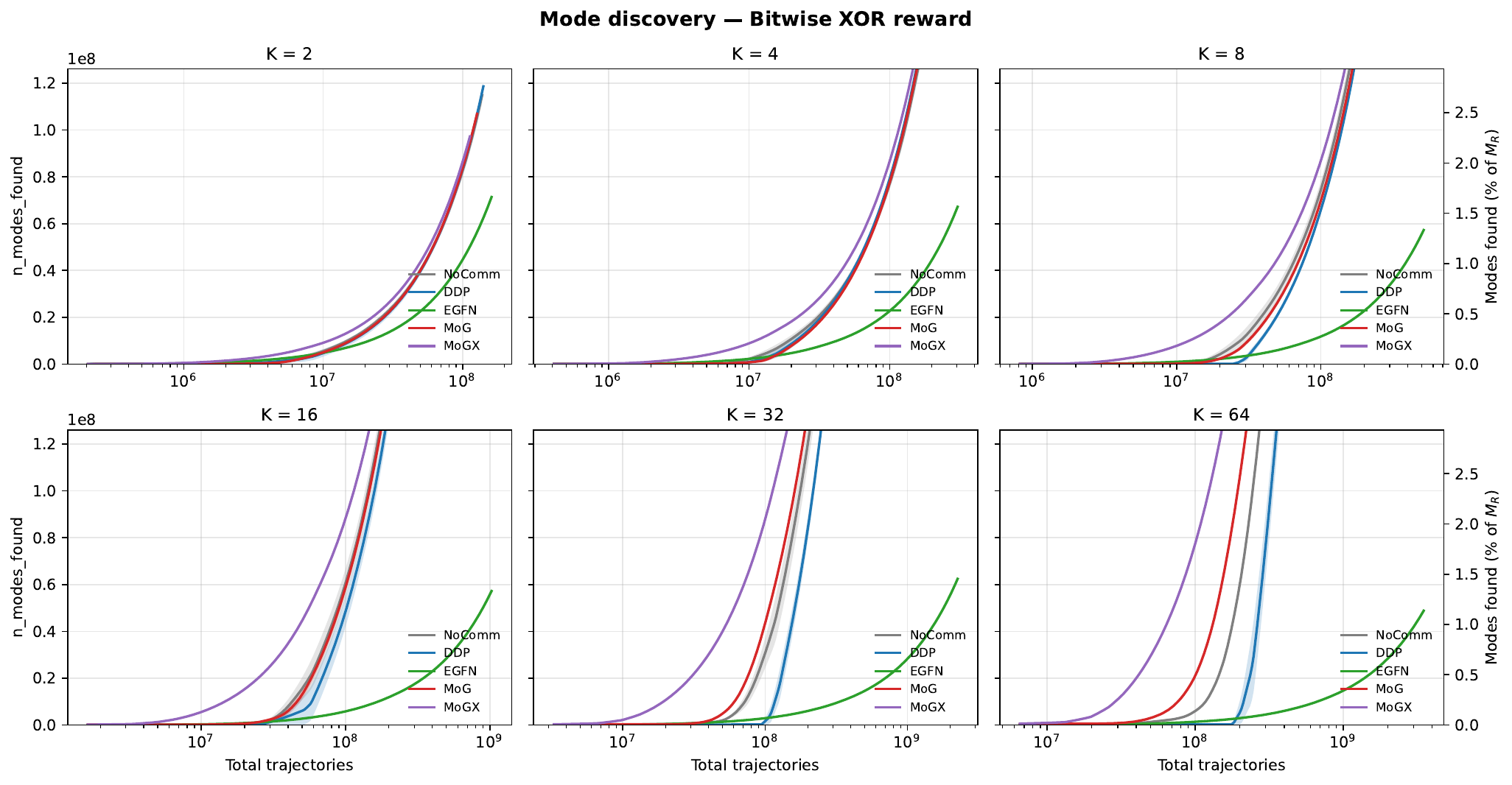}
    \caption{Modes discovered vs total trajectories on the
    \emph{bitwise\_xor} reward, \textbf{replay regime} ($\rho{=}0.5$).
    The MoGX--MoG gap widens with structural complexity.}
    \label{fig:lc_bitwise_xor}
\end{figure}

\begin{figure}[h]
    \centering
    \includegraphics[width=\linewidth]{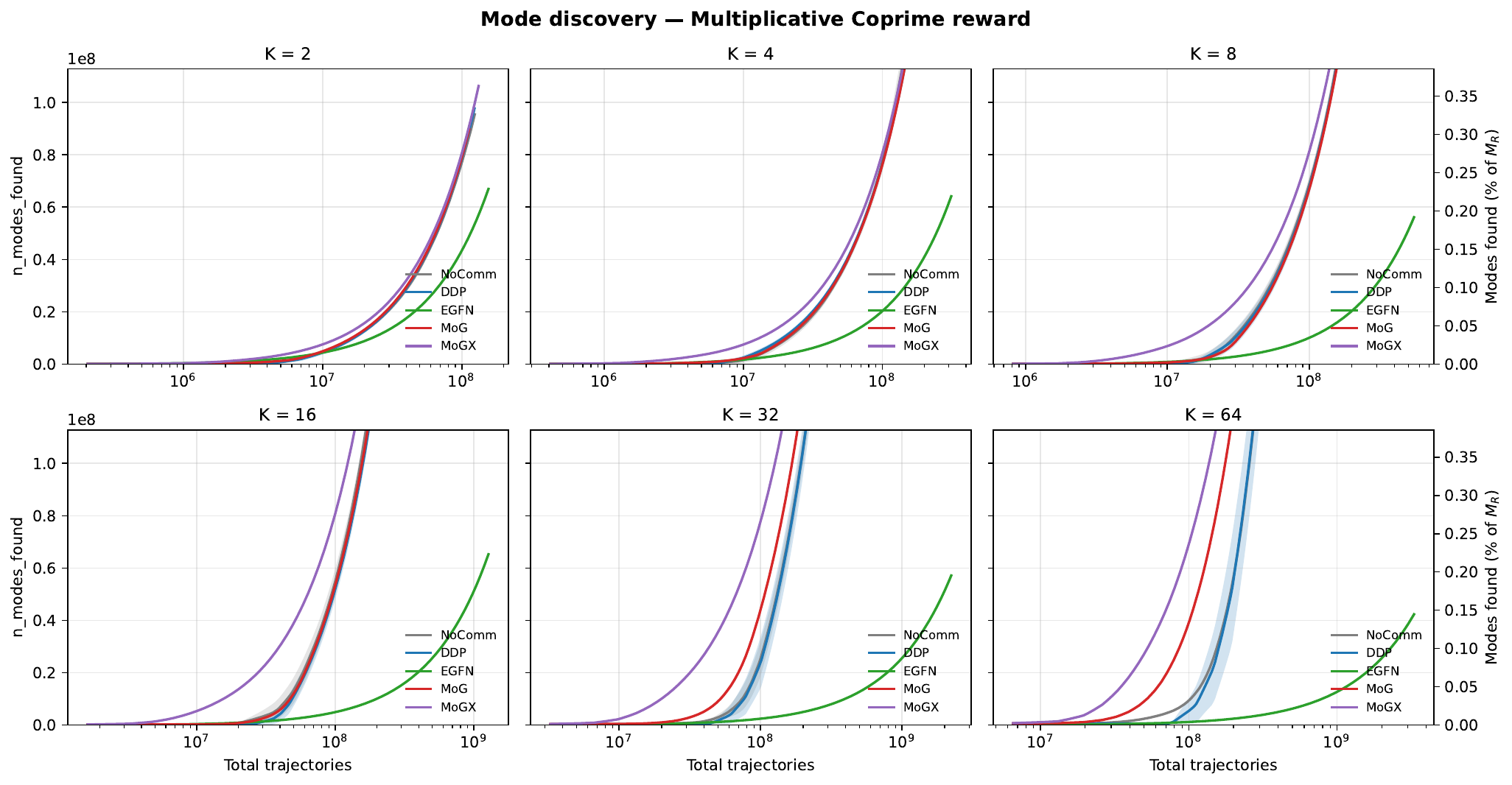}
    \caption{Modes discovered vs total trajectories on the
    \emph{multiplicative\_coprime} reward, \textbf{replay regime}
    ($\rho{=}0.5$). The MoGX--MoG gap widens with structural
    complexity.}
    \label{fig:lc_multiplicative_coprime}
\end{figure}

\begin{figure}[h]
    \centering
    \includegraphics[width=\linewidth]{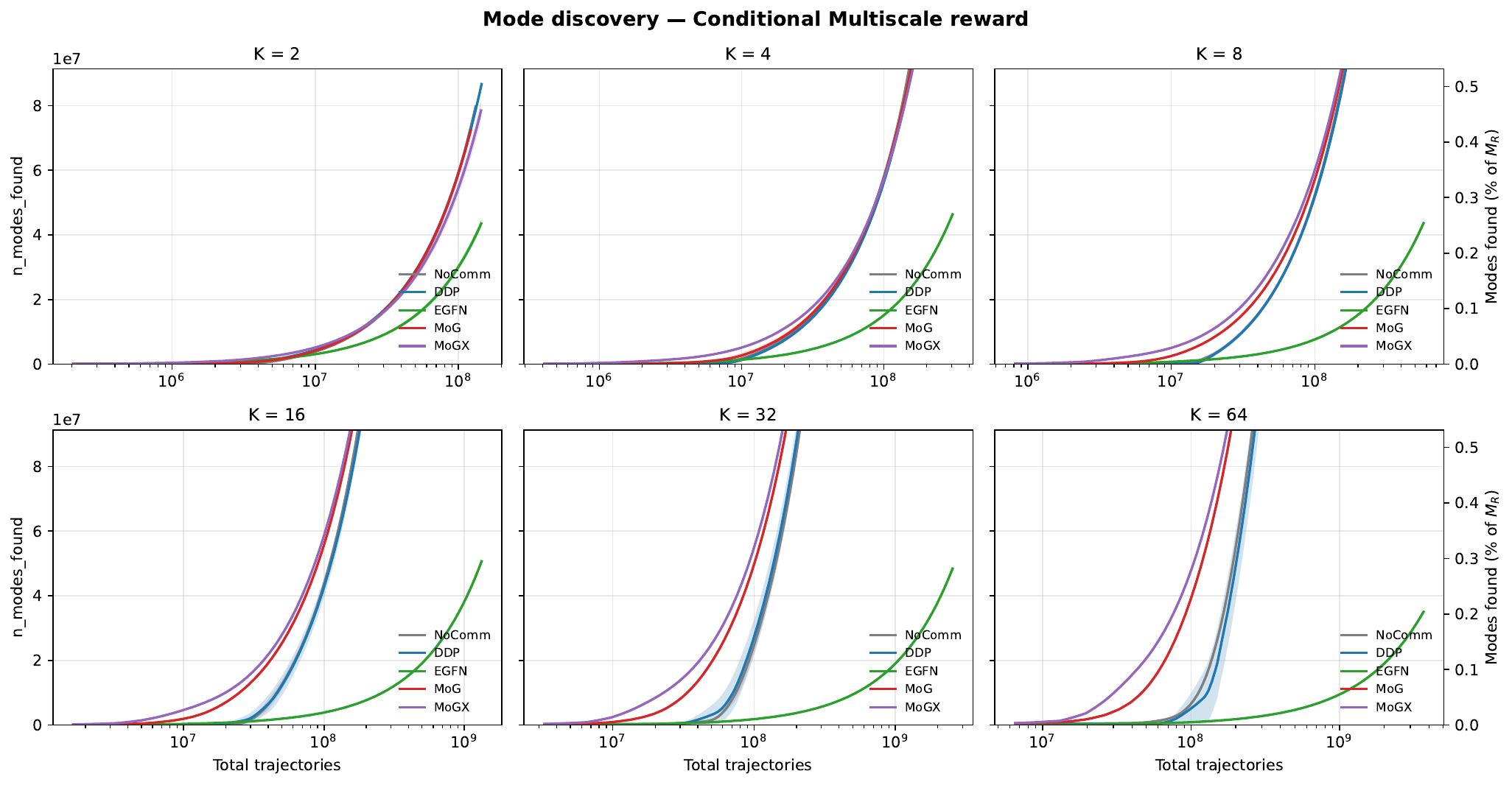}
    \caption{Modes discovered vs total trajectories on the
    \emph{conditional\_multiscale} reward, \textbf{replay regime}
    ($\rho{=}0.5$). The MoGX--MoG gap widens with structural
    complexity.}
    \label{fig:lc_conditional_multiscale}
\end{figure}

\begin{figure}[h]
    \centering
    \includegraphics[width=\linewidth]{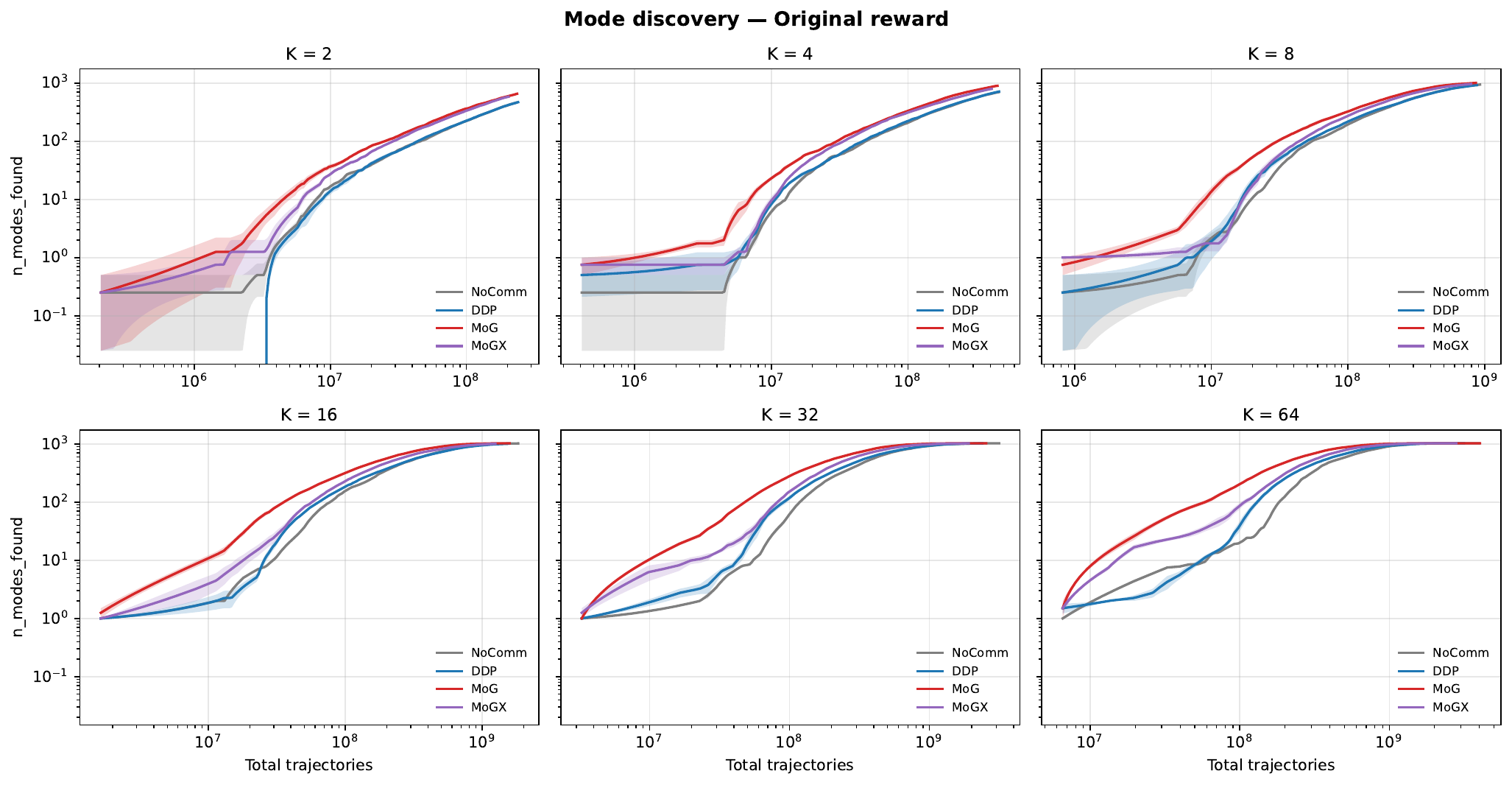}
    \caption{Modes discovered vs total trajectories on the
    \emph{original} reward, \textbf{pure on-policy regime} ($\rho{=}0$,
    replay buffer record-only).}
    \label{fig:lc_original_online}
\end{figure}

\begin{figure}[h]
    \centering
    \includegraphics[width=\linewidth]{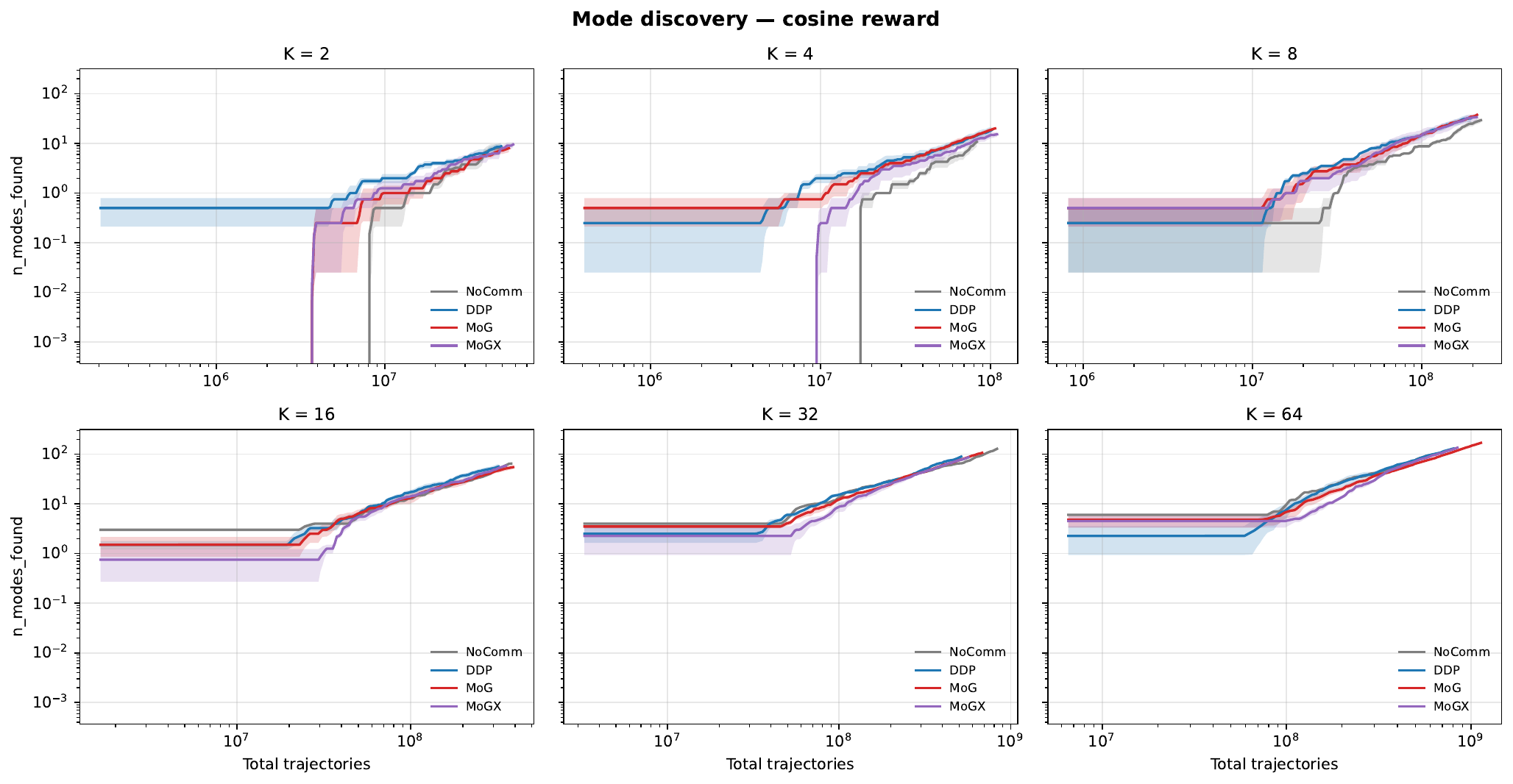}
    \caption{Modes discovered vs total trajectories on the \emph{cosine}
    reward, \textbf{pure on-policy regime} ($\rho{=}0$, replay buffer
    record-only).}
    \label{fig:lc_cosine_online}
\end{figure}

\begin{figure}[h]
    \centering
    \includegraphics[width=\linewidth]{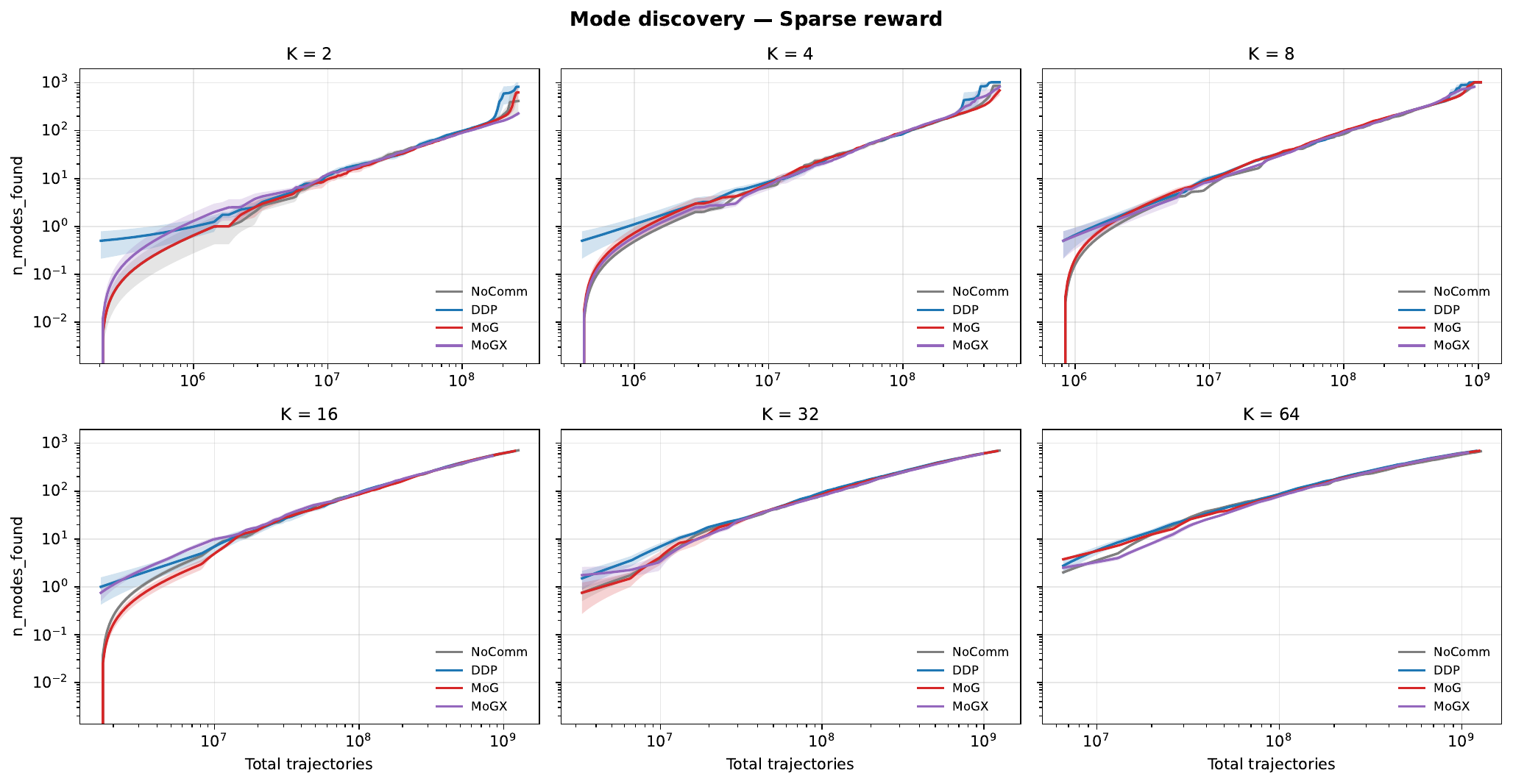}
    \caption{Modes discovered vs total trajectories on the \emph{sparse}
    reward, \textbf{pure on-policy regime} ($\rho{=}0$, replay buffer
    record-only).}
    \label{fig:lc_sparse_online}
\end{figure}

\begin{figure}[h]
    \centering
    \includegraphics[width=\linewidth]{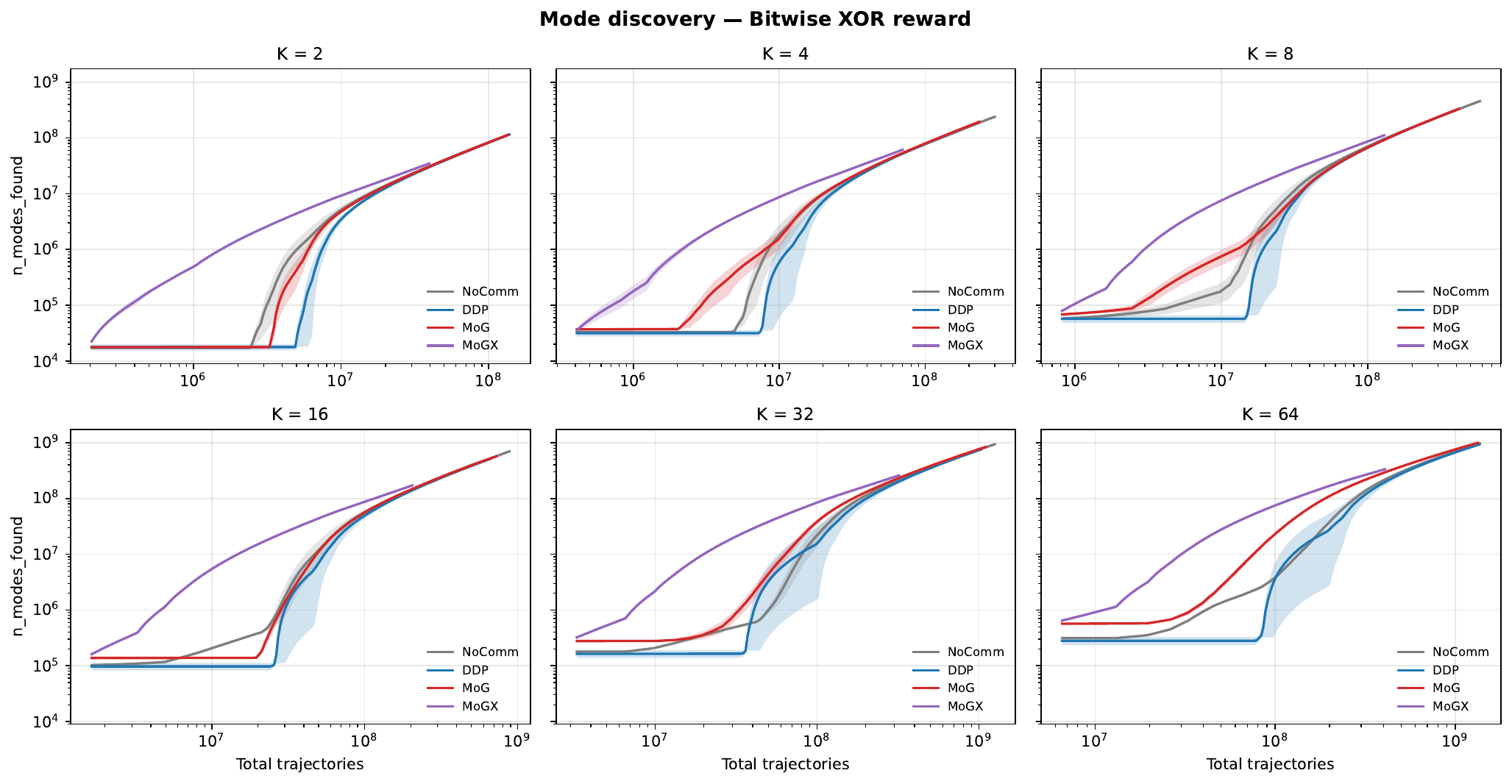}
    \caption{Modes discovered vs total trajectories on the
    \emph{bitwise\_xor} reward, \textbf{pure on-policy regime}
    ($\rho{=}0$, replay buffer record-only).}
    \label{fig:lc_bitwise_xor_online}
\end{figure}

\begin{figure}[h]
    \centering
    \includegraphics[width=\linewidth]{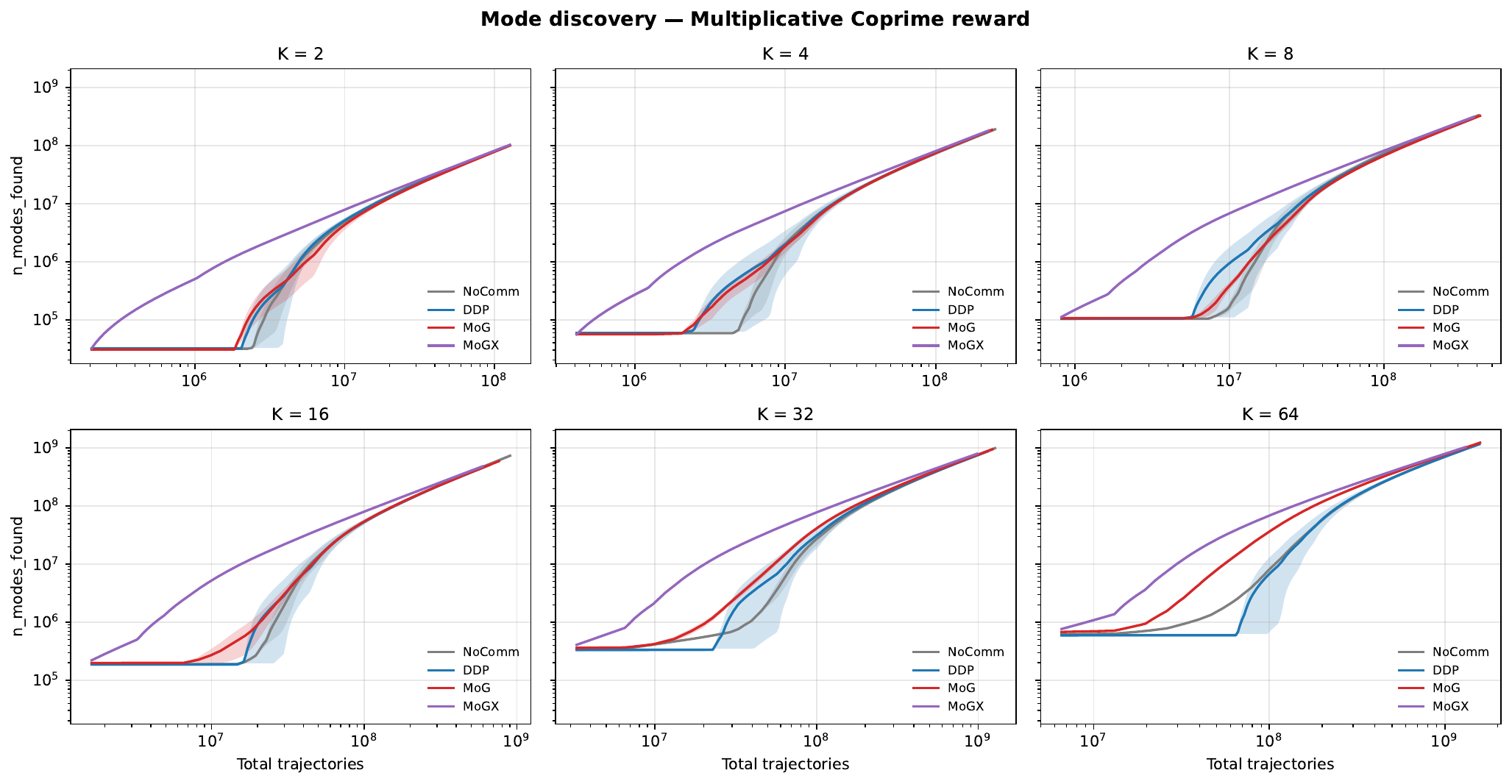}
    \caption{Modes discovered vs total trajectories on the
    \emph{multiplicative\_coprime} reward, \textbf{pure on-policy
    regime} ($\rho{=}0$, replay buffer record-only).}
    \label{fig:lc_multiplicative_coprime_online}
\end{figure}

\begin{figure}[h]
    \centering
    \includegraphics[width=\linewidth]{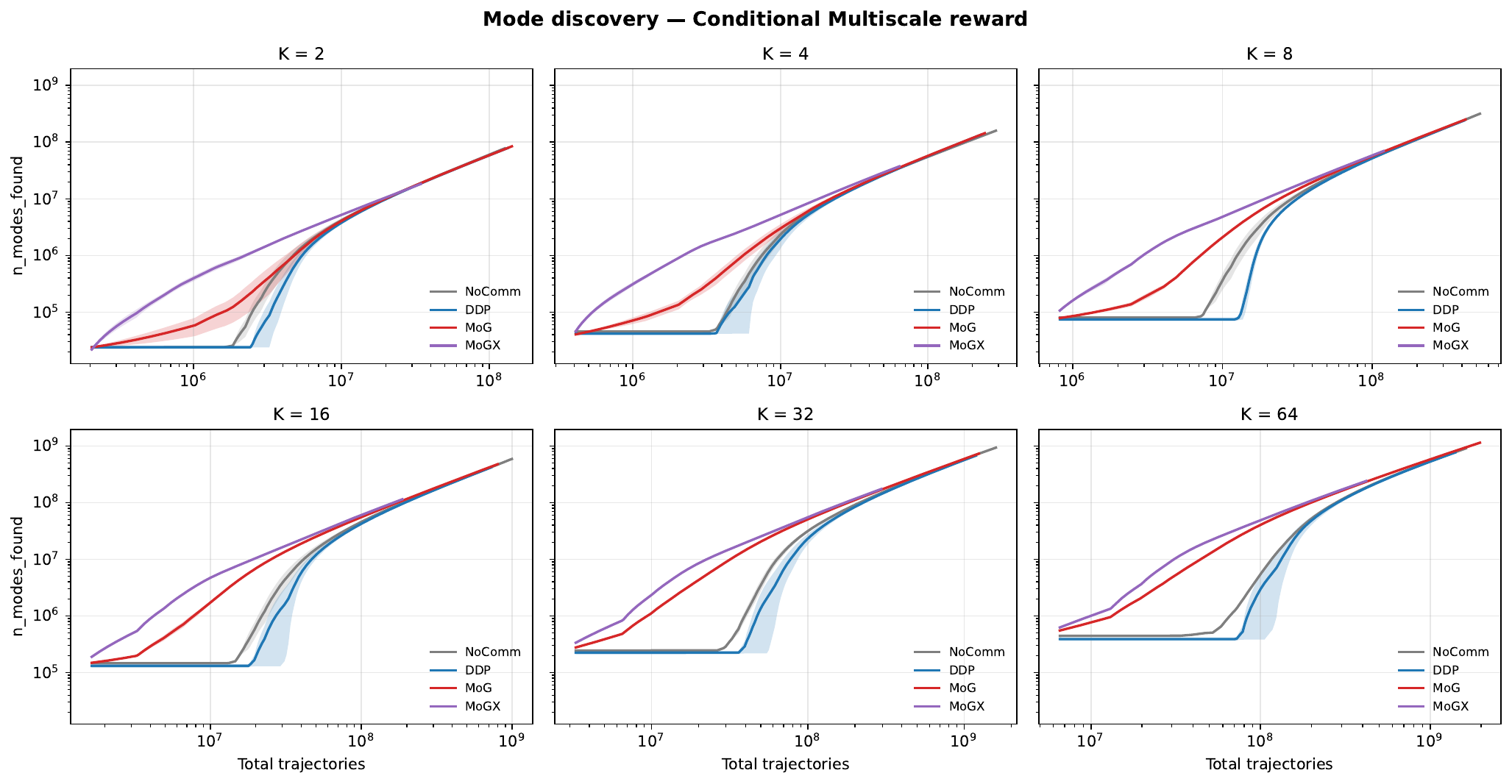}
    \caption{Modes discovered vs total trajectories on the
    \emph{conditional\_multiscale} reward, \textbf{pure on-policy
    regime} ($\rho{=}0$, replay buffer record-only).}
    \label{fig:lc_conditional_multiscale_online}
\end{figure}

\begin{figure}[h]
    \centering
    \includegraphics[width=\linewidth]{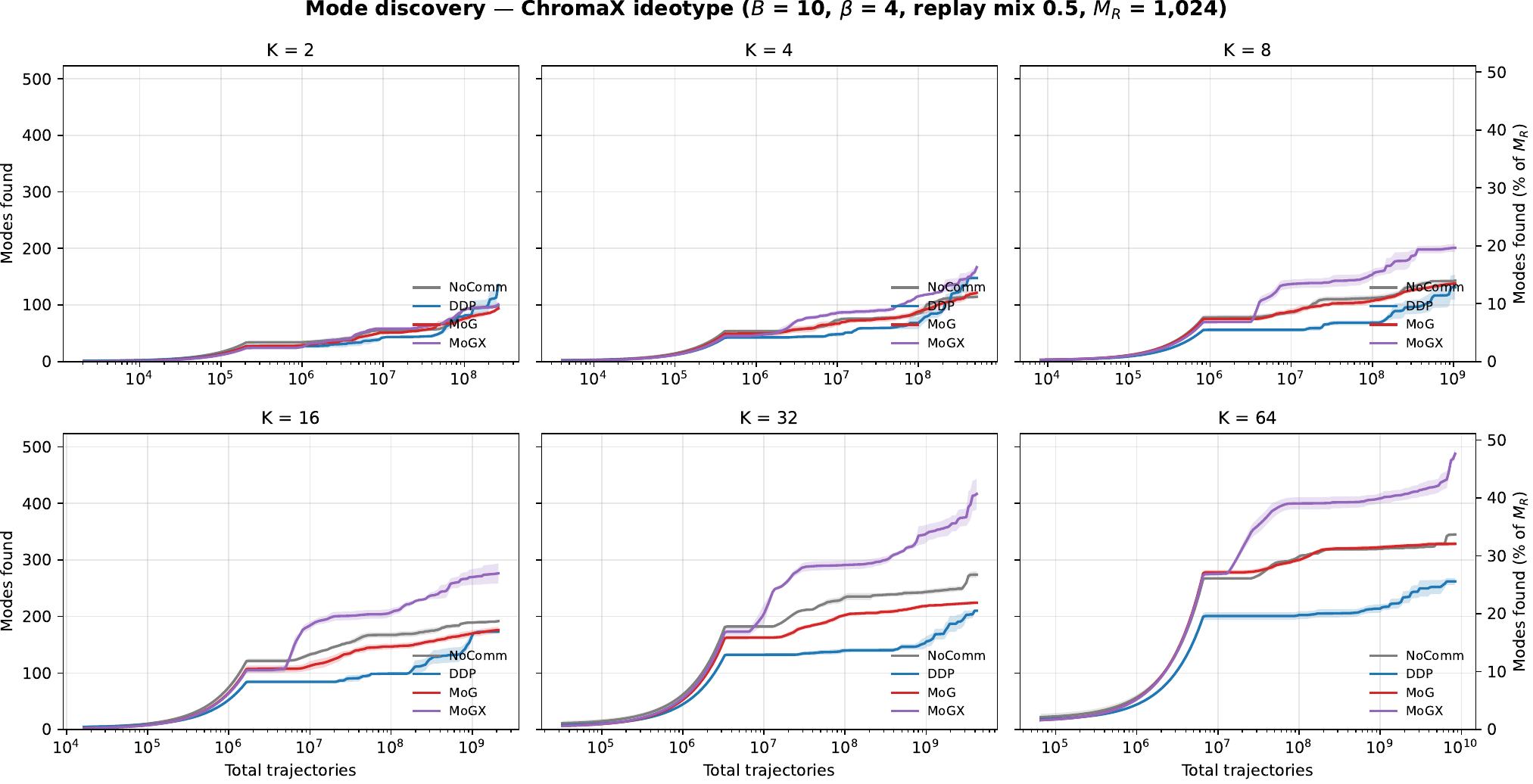}
    \caption{Modes discovered vs total trajectories on the ChromaX ideotype environment (\cref{sec:chromax}; $B{=}10$, $\beta{=}4$, $M_R{=}1024$ optima), \textbf{replay regime} ($\rho{=}0.5$), one panel per community size $K$, mean $\pm$ SEM over $4$ seeds. MoGX (purple) pulls ahead of the other three methods from $K{=}4$ on, while MoG (red) and NoComm (grey) stay close together throughout.}
    \label{fig:lc_chromax}
\end{figure}

\begin{figure}[h]
    \centering
    \includegraphics[width=\linewidth]{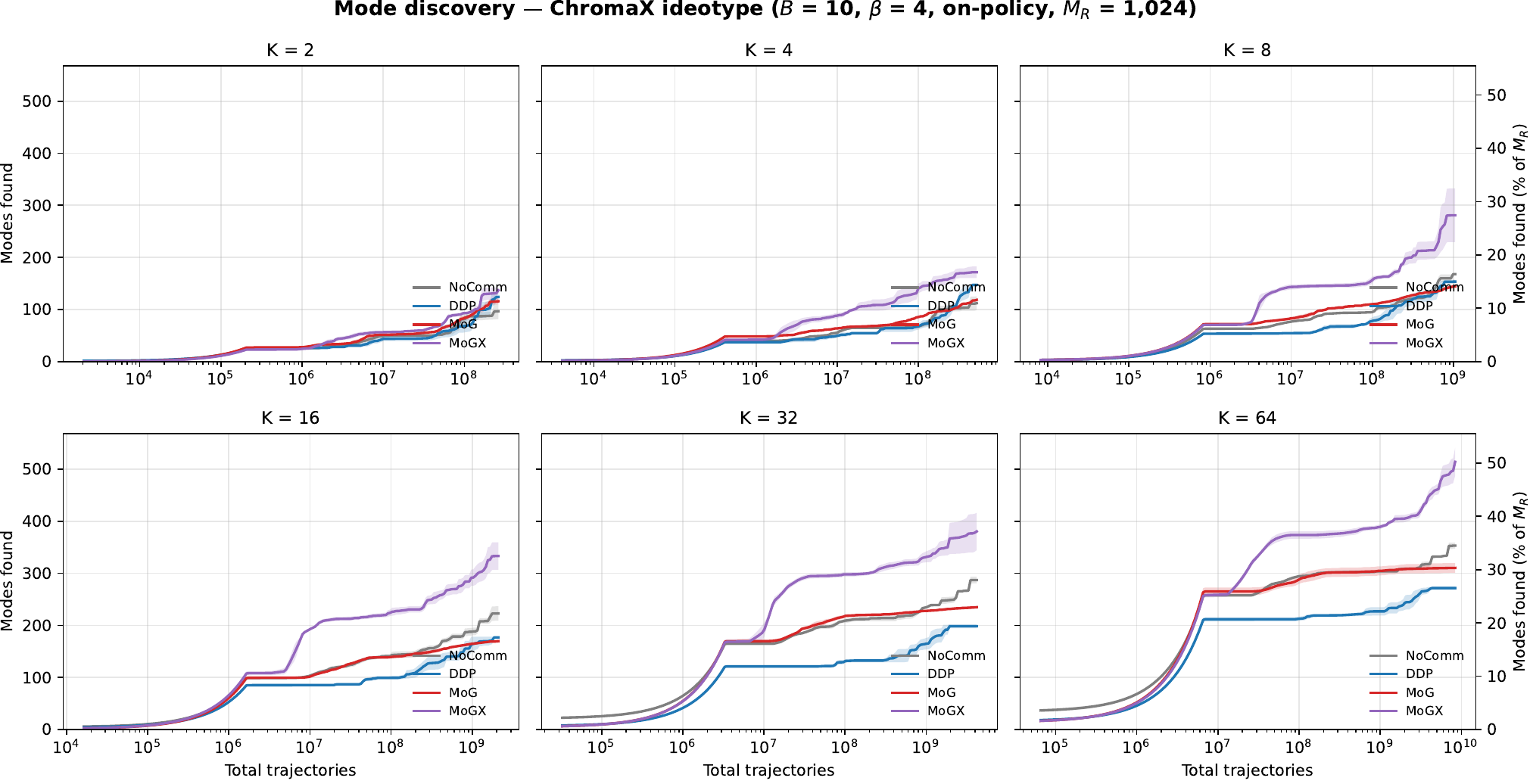}
    \caption{Modes discovered vs total trajectories on the ChromaX ideotype environment (\cref{sec:chromax}; $B{=}10$, $\beta{=}4$, $M_R{=}1024$ optima), \textbf{pure on-policy regime} ($\rho{=}0$, replay buffer record-only), one panel per community size $K$, mean $\pm$ SEM over $4$ seeds. The picture is the one from \cref{fig:lc_chromax}: MoGX (purple) separates from the other three methods from $K{=}4$ on, and MoG (red)
    stays on top of NoComm (grey).}
    \label{fig:lc_chromax_online}
\end{figure}

For completeness, we give the full per-trajectory mode-discovery curves
underlying the iso-budget endpoints reported in \cref{sec:experiments}.
\crefrange{fig:lc_original}{fig:lc_conditional_multiscale} cover the
replay regime ($\rho{=}0.5$) and
\crefrange{fig:lc_original_online}{fig:lc_conditional_multiscale_online}
the pure on-policy regime ($\rho{=}0$), with one figure per reward. In
general, the curves show that
MoG and MoGX are more sample efficient than NoComm and DDP and continue benefiting from larger $K$. In each figure, columns subdivide by community size $K$ and
shaded bands show $\pm 1$ s.d.\ across seeds.
\Cref{fig:lc_chromax,fig:lc_chromax_online}
give the same view for the ChromaX ideotype environment of \cref{sec:chromax},
in the replay and pure on-policy regimes respectively.

% ---- Replay regime ($\rho=0.5$), non-tiered rewards ----

% ---- Replay regime ($\rho=0.5$), tiered/compositional rewards ----

% ---- Pure on-policy regime ($\rho=0$), non-tiered rewards ----

% ---- Pure on-policy regime ($\rho=0$), tiered/compositional rewards ----

% ---- ChromaX ideotype environment ----

\section{Other Training Objectives}
\label{app:objectives}

We reran the pure on-policy sweep ($\rho{=}0$) with SubTB and DB on
\emph{original} (easy preset, $R_0{=}0.01$), \emph{bitwise\_xor} and
\emph{conditional\_multiscale}. SubTB uses $\lambda{=}0.9$ with geometric
weighting within each sub-trajectory, and for both losses the $\log F$ head
shares the policy trunk. Everything else follows \cref{tab:hyperparams}.
Endpoints are read at the largest budget every run with $K{\geq}2$ reached,
over 4 seeds.

\begin{figure}[h]
    \centering
    \includegraphics[width=\linewidth]{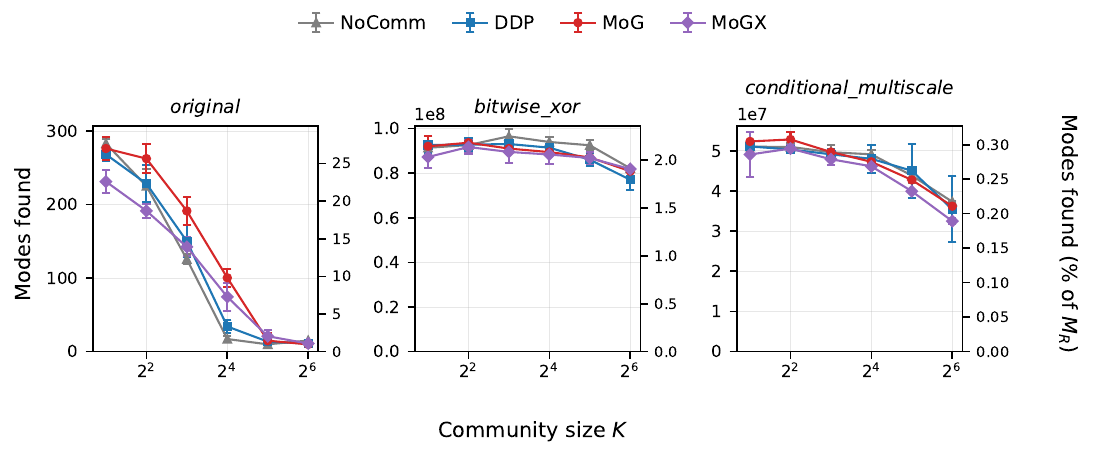}
    \caption{Modes discovered at fixed total-trajectory budget vs community
    size $K$, \textbf{SubTB}, pure on-policy regime, mean $\pm 1$ s.d.\ over
    4 seeds.}
    \label{fig:objectives_subtb}
\end{figure}

\paragraph{SubTB.}
SubTB's baselines barely suffer from a larger community. On
\emph{bitwise\_xor}, NoComm finds $9.1\times10^{7}$ modes at $K{=}2$ and still
$8.2\times10^{7}$ at $K{=}64$, where all four methods sit within $7\%$ of each
other (\cref{fig:objectives_subtb}). There is no collapse for MoG to prevent.
The one clear gain is on \emph{original} at intermediate $K$: at $K{=}16$ MoG
finds $100$ modes, DDP $34$ and NoComm $17$. So the TB gains of
\cref{fig:final_modes_vs_K_online} depend on how badly TB's baselines degrade
with $K$.

\paragraph{DB.}
Under DB, MoG and MoGX carry the tilt as a flow potential. Rank $k$ uses
\[
\log F(s) \;=\; \log F_\theta(s) + u_k(s),
\qquad
u_k(s) \;=\; \max\bigl(\log f_\psi(s)_k,\, -5\bigr),
\]
and shapes terminal rewards with the same $u_k$. The optimum is that of DB
with boundary $\log\tilde R_k$, but every interior residual now carries
$u_k(s)-u_k(s')$ from the first update, and $\log F_\theta$ only fits the
untilted flow.

\begin{figure}[h]
    \centering
    \includegraphics[width=\linewidth]{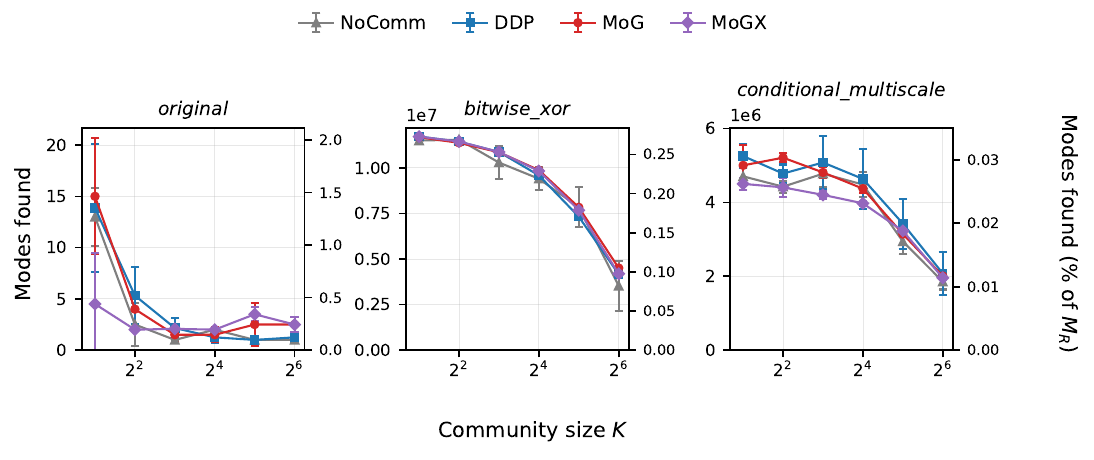}
    \caption{Same as \cref{fig:objectives_subtb} for \textbf{DB}, with the
    tilt carried as a flow potential in MoG and MoGX.}
    \label{fig:objectives_db}
\end{figure}

MoG tracks the baselines up to $K{=}16$ and pulls ahead on
\emph{bitwise\_xor} beyond it, with $6.5\times10^{7}$ modes at $K{=}64$
against $6.2\times10^{7}$ for DDP and $6.0\times10^{7}$ for NoComm
(\cref{fig:objectives_db}). On the other two rewards, it shows no consistent
gain. MoGX is worse: it trails every method at every $K$ on
\emph{bitwise\_xor}, and at $K{\leq}16$ on \emph{original} it finds $17$--$51$
modes where the others find $95$--$229$.

\section{Timings}
\label{app:timings}

\paragraph{Per-reward totals.}
\Cref{tab:timings_total} of the main paper reports the per-method
per-iteration total for the \emph{original} reward.
\Cref{tab:timings_total_original,tab:timings_total_cosine,tab:timings_total_sparse,tab:timings_total_bitwise_xor,tab:timings_total_multiplicative_coprime,tab:timings_total_conditional_multiscale}
give the same total for each reward separately, with \cref{tab:timings_total_original}
reproducing the main-paper table. The per-method scaling pattern is identical for every reward:
NoComm stays flat in $K$; DDP grows with the $\nabla_\theta$
all-reduce; MoG grows much more slowly, paying only the much cheaper
classifier all-reduce; and MoGX adds the trajectory all-to-all on top.
The absolute level shifts with the per-reward shaping cost
(\emph{sparse} is the cheapest, \emph{cosine} the most expensive). At $K{=}64$
the ordering NoComm $<$ MoG $<$ DDP $<$ MoGX holds on every reward except \emph{sparse},
where MoG and NoComm are within noise. At small $K$ the methods are within run-to-run
noise of each other and the ordering varies. NoComm, whose per-rank work does not
depend on $K$, varies by up to $23\%$ across $K$ on a single reward.

\begin{table}[h]
\centering
\footnotesize
\setlength{\tabcolsep}{4pt}
\begin{tabular}{lrrrrrrr}
\toprule
\textbf{Method} & \textbf{$K{=}1$} & \textbf{$2$} & \textbf{$4$} & \textbf{$8$} & \textbf{$16$} & \textbf{$32$} & \textbf{$64$} \\
\midrule
\textbf{NoComm} & 702 \SE{4} & 729 \SE{6} & 726 \SE{7} & 728 \SE{7} & 732 \SE{3} & 758 \SE{23} & 713 \SE{9} \\
\textbf{DDP} & 702 \SE{4} & 715 \SE{3} & 743 \SE{2} & 784 \SE{1} & 823 \SE{2} & 950 \SE{4} & 1249 \SE{11} \\
\textbf{EGFN} & 702 \SE{4} & 712 \SE{3} & 748 \SE{23} & 853 \SE{14} & 876 \SE{70} & 803 \SE{6} & 975 \SE{3} \\
\textbf{MoG} & 702 \SE{4} & 721 \SE{4} & 749 \SE{7} & 800 \SE{3} & 818 \SE{6} & 923 \SE{8} & 976 \SE{7} \\
\textbf{MoGX} & 702 \SE{4} & 762 \SE{30} & 804 \SE{15} & 841 \SE{12} & 998 \SE{12} & 1215 \SE{49} & 1477 \SE{12} \\
\bottomrule
\end{tabular}
\caption{Same per-method per-iteration wall-clock total as \cref{tab:timings_total}, reported also here for reference.}
\label{tab:timings_total_original}
\end{table}

\begin{table}[h]
\centering
\footnotesize
\setlength{\tabcolsep}{4pt}
\begin{tabular}{lrrrrrrr}
\toprule
\textbf{Method} & \textbf{$K{=}1$} & \textbf{$2$} & \textbf{$4$} & \textbf{$8$} & \textbf{$16$} & \textbf{$32$} & \textbf{$64$} \\
\midrule
\textbf{NoComm} & 2936 \SE{45} & 3103 \SE{167} & 3278 \SE{278} & 2971 \SE{65} & 3303 \SE{122} & 2999 \SE{71} & 2969 \SE{73} \\
\textbf{DDP} & 2936 \SE{45} & 3106 \SE{142} & 3127 \SE{94} & 3181 \SE{123} & 3451 \SE{224} & 3988 \SE{204} & 4727 \SE{182} \\
\textbf{MoG} & 2936 \SE{45} & 2982 \SE{68} & 3099 \SE{52} & 3197 \SE{26} & 3274 \SE{37} & 3628 \SE{40} & 3994 \SE{31} \\
\textbf{MoGX} & 2936 \SE{45} & 2919 \SE{24} & 3092 \SE{35} & 3218 \SE{15} & 3643 \SE{64} & 4059 \SE{34} & 4840 \SE{17} \\
\bottomrule
\end{tabular}
\caption{Per-method per-iteration wall-clock total, similar to \cref{tab:timings_total}, but restricted to the \emph{cosine} reward. Aggregation, units, and method definitions match \cref{tab:timings_total}. At $K{=}1$ all four methods collapse to the same single-agent loop and share the pooled K=1 baseline.}
\label{tab:timings_total_cosine}
\end{table}

\begin{table}[h]
\centering
\footnotesize
\setlength{\tabcolsep}{4pt}
\begin{tabular}{lrrrrrrr}
\toprule
\textbf{Method} & \textbf{$K{=}1$} & \textbf{$2$} & \textbf{$4$} & \textbf{$8$} & \textbf{$16$} & \textbf{$32$} & \textbf{$64$} \\
\midrule
\textbf{NoComm} & 494 \SE{3} & 498 \SE{2} & 479 \SE{5} & 479 \SE{5} & 464 \SE{3} & 474 \SE{5} & 483 \SE{4} \\
\textbf{DDP} & 494 \SE{3} & 529 \SE{3} & 560 \SE{3} & 606 \SE{3} & 905 \SE{21} & 1475 \SE{48} & 2780 \SE{120} \\
\textbf{MoG} & 494 \SE{3} & 501 \SE{9} & 522 \SE{6} & 558 \SE{3} & 526 \SE{9} & 488 \SE{7} & 476 \SE{13} \\
\textbf{MoGX} & 494 \SE{3} & 556 \SE{8} & 601 \SE{8} & 701 \SE{4} & 1519 \SE{25} & 2640 \SE{46} & 4535 \SE{53} \\
\bottomrule
\end{tabular}
\caption{Per-method per-iteration wall-clock total, similar to \cref{tab:timings_total}, but restricted to the \emph{sparse} reward. Aggregation, units, and method definitions match \cref{tab:timings_total}. At $K{=}1$ all four methods collapse to the same single-agent loop and share the pooled K=1 baseline.}
\label{tab:timings_total_sparse}
\end{table}

\begin{table}[h]
\centering
\footnotesize
\setlength{\tabcolsep}{4pt}
\begin{tabular}{lrrrrrrr}
\toprule
\textbf{Method} & \textbf{$K{=}1$} & \textbf{$2$} & \textbf{$4$} & \textbf{$8$} & \textbf{$16$} & \textbf{$32$} & \textbf{$64$} \\
\midrule
\textbf{NoComm} & 1083 \SE{22} & 1151 \SE{48} & 1061 \SE{26} & 1066 \SE{22} & 1094 \SE{47} & 1067 \SE{9} & 937 \SE{23} \\
\textbf{DDP} & 1083 \SE{22} & 1206 \SE{19} & 1402 \SE{43} & 1636 \SE{25} & 1829 \SE{29} & 1904 \SE{41} & 2530 \SE{81} \\
\textbf{MoG} & 1083 \SE{22} & 1208 \SE{39} & 1363 \SE{40} & 1567 \SE{22} & 1543 \SE{15} & 1578 \SE{16} & 1470 \SE{44} \\
\textbf{MoGX} & 1083 \SE{22} & 1199 \SE{59} & 1388 \SE{89} & 1619 \SE{130} & 1901 \SE{147} & 2291 \SE{171} & 2923 \SE{215} \\
\bottomrule
\end{tabular}
\caption{Per-method per-iteration wall-clock total, similar to \cref{tab:timings_total}, but restricted to the \emph{bitwise\_xor} reward. Aggregation, units, and method definitions match \cref{tab:timings_total}. At $K{=}1$ all four methods collapse to the same single-agent loop and share the pooled K=1 baseline.}
\label{tab:timings_total_bitwise_xor}
\end{table}

\begin{table}[h]
\centering
\footnotesize
\setlength{\tabcolsep}{4pt}
\begin{tabular}{lrrrrrrr}
\toprule
\textbf{Method} & \textbf{$K{=}1$} & \textbf{$2$} & \textbf{$4$} & \textbf{$8$} & \textbf{$16$} & \textbf{$32$} & \textbf{$64$} \\
\midrule
\textbf{NoComm} & 1162 \SE{25} & 1320 \SE{98} & 1238 \SE{60} & 1328 \SE{106} & 1319 \SE{43} & 1216 \SE{63} & 1102 \SE{53} \\
\textbf{DDP} & 1162 \SE{25} & 1348 \SE{40} & 1495 \SE{28} & 1725 \SE{30} & 1906 \SE{33} & 1945 \SE{15} & 2315 \SE{40} \\
\textbf{MoG} & 1162 \SE{25} & 1244 \SE{41} & 1415 \SE{21} & 1598 \SE{27} & 1615 \SE{17} & 1789 \SE{30} & 1671 \SE{20} \\
\textbf{MoGX} & 1162 \SE{25} & 1325 \SE{17} & 1505 \SE{11} & 1774 \SE{20} & 2086 \SE{74} & 2300 \SE{26} & 2795 \SE{48} \\
\bottomrule
\end{tabular}
\caption{Per-method per-iteration wall-clock total, similar to \cref{tab:timings_total}, but restricted to the \emph{multiplicative\_coprime} reward. Aggregation, units, and method definitions match \cref{tab:timings_total}. At $K{=}1$ all four methods collapse to the same single-agent loop and share the pooled K=1 baseline.}
\label{tab:timings_total_multiplicative_coprime}
\end{table}

\begin{table}[h]
\centering
\footnotesize
\setlength{\tabcolsep}{4pt}
\begin{tabular}{lrrrrrrr}
\toprule
\textbf{Method} & \textbf{$K{=}1$} & \textbf{$2$} & \textbf{$4$} & \textbf{$8$} & \textbf{$16$} & \textbf{$32$} & \textbf{$64$} \\
\midrule
\textbf{NoComm} & 1114 \SE{24} & 1272 \SE{45} & 1110 \SE{26} & 1220 \SE{33} & 1174 \SE{39} & 1171 \SE{65} & 1087 \SE{45} \\
\textbf{DDP} & 1114 \SE{24} & 1241 \SE{31} & 1433 \SE{36} & 1597 \SE{30} & 1736 \SE{20} & 1758 \SE{33} & 2329 \SE{136} \\
\textbf{MoG} & 1114 \SE{24} & 1180 \SE{16} & 1323 \SE{37} & 1518 \SE{37} & 1580 \SE{12} & 1674 \SE{17} & 1672 \SE{38} \\
\textbf{MoGX} & 1114 \SE{24} & 1194 \SE{72} & 1364 \SE{99} & 1523 \SE{112} & 1841 \SE{165} & 2020 \SE{171} & 2697 \SE{265} \\
\bottomrule
\end{tabular}
\caption{Per-method per-iteration wall-clock total, similar to \cref{tab:timings_total}, but restricted to the \emph{conditional\_multiscale} reward. Aggregation, units, and method definitions match \cref{tab:timings_total}. At $K{=}1$ all four methods collapse to the same single-agent loop and share the pooled K=1 baseline.}
\label{tab:timings_total_conditional_multiscale}
\end{table}

\paragraph{Per-reward phase breakdown.}
We also time the different training phases while scaling the community size. \Cref{tab:timings_original,tab:timings_cosine,tab:timings_sparse,tab:timings_bitwise_xor,tab:timings_multiplicative_coprime,tab:timings_conditional_multiscale}
split the measurement by reward. \emph{Sample} = on-policy rollouts; \emph{Loss} = reward
shaping + GFN loss forward; \emph{Bwd} = backward pass. The rank-local phases are pooled over
all methods. The cross-rank phases are averaged over the methods that use them:
\emph{Exch.}$^\dagger$ = MoGX trajectory all-to-all; $\nabla_\theta^{\ddagger}$ = DDP
policy-gradient all-reduce; $f_\psi^{\star}$ = classifier loss + blocking gradient all-reduce,
averaged over MoG and MoGX. The two differ: on \emph{original}, MoG's $f_\psi$ phase takes
$23$\,ms at $K{=}2$ and $284$\,ms at $K{=}64$, while MoGX's stays below $12$\,ms.
The rank-local phases (Sample, Loss, Bwd)
track the per-reward shaping cost: \emph{sparse} is the cheapest, while
\emph{cosine} is the most expensive. The cross-rank phases (Exch.,
$\nabla_\theta$, $f_\psi$) follow the same qualitative scaling in $K$
as in the aggregate, independent of the reward.

\begin{table}[h]
\centering
\footnotesize
\setlength{\tabcolsep}{6pt}
\begin{tabular}{rrrrrrr}
\toprule
\textbf{$K$} & \textbf{Sample} & \textbf{Loss} & \textbf{Bwd} & \textbf{Exch.$^\dagger$} & \textbf{$\nabla_\theta^{\ddagger}$} & \textbf{$f_\psi^{\star}$} \\
\midrule
\textbf{1} & 246 \SE{1} & 224 \SE{1} & 232 \SE{2} & -- & -- & -- \\
\textbf{2} & 245 \SE{2} & 224 \SE{2} & 229 \SE{3} & 84 \SE{45} & 24 \SE{3} & 14 \SE{4} \\
\textbf{4} & 245 \SE{3} & 227 \SE{3} & 233 \SE{4} & 92 \SE{44} & 55 \SE{2} & 28 \SE{14} \\
\textbf{8} & 248 \SE{4} & 230 \SE{4} & 235 \SE{4} & 101 \SE{29} & 90 \SE{3} & 54 \SE{22} \\
\textbf{16} & 244 \SE{3} & 233 \SE{4} & 238 \SE{4} & 263 \SE{42} & 129 \SE{4} & 60 \SE{22} \\
\textbf{32} & 244 \SE{3} & 232 \SE{6} & 238 \SE{7} & 514 \SE{102} & 256 \SE{7} & 108 \SE{37} \\
\textbf{64} & 245 \SE{2} & 226 \SE{3} & 233 \SE{3} & 760 \SE{38} & 549 \SE{11} & 145 \SE{52} \\
\bottomrule
\end{tabular}
\caption{Per-iteration timing phase breakdown, restricted to the \emph{original} reward.}
\label{tab:timings_original}
\end{table}

\begin{table}[h]
\centering
\footnotesize
\setlength{\tabcolsep}{6pt}
\begin{tabular}{rrrrrrr}
\toprule
\textbf{$K$} & \textbf{Sample} & \textbf{Loss} & \textbf{Bwd} & \textbf{Exch.$^\dagger$} & \textbf{$\nabla_\theta^{\ddagger}$} & \textbf{$f_\psi^{\star}$} \\
\midrule
\textbf{1} & 914 \SE{7} & 1051 \SE{26} & 971 \SE{20} & -- & -- & -- \\
\textbf{2} & 915 \SE{11} & 1052 \SE{27} & 971 \SE{22} & 88 \SE{56} & 231 \SE{153} & 21 \SE{9} \\
\textbf{4} & 912 \SE{14} & 1087 \SE{50} & 993 \SE{36} & 312 \SE{106} & 210 \SE{56} & 53 \SE{22} \\
\textbf{8} & 909 \SE{12} & 1042 \SE{17} & 961 \SE{15} & 362 \SE{45} & 245 \SE{42} & 155 \SE{71} \\
\textbf{16} & 901 \SE{15} & 1095 \SE{39} & 995 \SE{32} & 773 \SE{217} & 464 \SE{129} & 233 \SE{98} \\
\textbf{32} & 895 \SE{13} & 1056 \SE{26} & 965 \SE{20} & 1220 \SE{60} & 983 \SE{64} & 403 \SE{161} \\
\textbf{64} & 913 \SE{16} & 1038 \SE{24} & 955 \SE{22} & 1981 \SE{122} & 1668 \SE{50} & 627 \SE{239} \\
\bottomrule
\end{tabular}
\caption{Per-iteration timing phase breakdown, restricted to the \emph{cosine} reward.}
\label{tab:timings_cosine}
\end{table}

\begin{table}[h]
\centering
\footnotesize
\setlength{\tabcolsep}{6pt}
\begin{tabular}{rrrrrrr}
\toprule
\textbf{$K$} & \textbf{Sample} & \textbf{Loss} & \textbf{Bwd} & \textbf{Exch.$^\dagger$} & \textbf{$\nabla_\theta^{\ddagger}$} & \textbf{$f_\psi^{\star}$} \\
\midrule
\textbf{1} & 170 \SE{1} & 167 \SE{1} & 157 \SE{1} & -- & -- & -- \\
\textbf{2} & 172 \SE{2} & 172 \SE{2} & 161 \SE{2} & 22 \SE{9} & 21 \SE{3} & 9 \SE{1} \\
\textbf{4} & 167 \SE{1} & 170 \SE{2} & 158 \SE{2} & 79 \SE{26} & 61 \SE{8} & 23 \SE{7} \\
\textbf{8} & 167 \SE{2} & 169 \SE{3} & 157 \SE{3} & 188 \SE{34} & 103 \SE{7} & 40 \SE{13} \\
\textbf{16} & 163 \SE{2} & 165 \SE{3} & 153 \SE{3} & 1032 \SE{50} & 416 \SE{25} & 20 \SE{6} \\
\textbf{32} & 162 \SE{2} & 164 \SE{2} & 152 \SE{2} & 2164 \SE{33} & 986 \SE{53} & 6 \SE{0} \\
\textbf{64} & 166 \SE{3} & 168 \SE{3} & 156 \SE{3} & 4017 \SE{77} & 2280 \SE{116} & 10 \SE{2} \\
\bottomrule
\end{tabular}
\caption{Per-iteration timing phase breakdown, restricted to the \emph{sparse} reward.}
\label{tab:timings_sparse}
\end{table}

\begin{table}[h]
\centering
\footnotesize
\setlength{\tabcolsep}{6pt}
\begin{tabular}{rrrrrrr}
\toprule
\textbf{$K$} & \textbf{Sample} & \textbf{Loss} & \textbf{Bwd} & \textbf{Exch.$^\dagger$} & \textbf{$\nabla_\theta^{\ddagger}$} & \textbf{$f_\psi^{\star}$} \\
\midrule
\textbf{1} & 435 \SE{7} & 325 \SE{10} & 323 \SE{10} & -- & -- & -- \\
\textbf{2} & 423 \SE{7} & 331 \SE{11} & 329 \SE{11} & 177 \SE{32} & 142 \SE{8} & 54 \SE{19} \\
\textbf{4} & 410 \SE{8} & 312 \SE{12} & 308 \SE{12} & 414 \SE{58} & 380 \SE{39} & 149 \SE{59} \\
\textbf{8} & 394 \SE{9} & 318 \SE{11} & 313 \SE{11} & 563 \SE{109} & 608 \SE{31} & 307 \SE{113} \\
\textbf{16} & 373 \SE{9} & 307 \SE{8} & 304 \SE{9} & 1006 \SE{134} & 832 \SE{19} & 295 \SE{109} \\
\textbf{32} & 346 \SE{11} & 312 \SE{12} & 308 \SE{11} & 1455 \SE{131} & 964 \SE{36} & 279 \SE{104} \\
\textbf{64} & 323 \SE{7} & 300 \SE{11} & 296 \SE{11} & 2013 \SE{125} & 1624 \SE{75} & 272 \SE{106} \\
\bottomrule
\end{tabular}
\caption{Per-iteration timing phase breakdown, restricted to the \emph{bitwise\_xor} reward.}
\label{tab:timings_bitwise_xor}
\end{table}

\begin{table}[h]
\centering
\footnotesize
\setlength{\tabcolsep}{6pt}
\begin{tabular}{rrrrrrr}
\toprule
\textbf{$K$} & \textbf{Sample} & \textbf{Loss} & \textbf{Bwd} & \textbf{Exch.$^\dagger$} & \textbf{$\nabla_\theta^{\ddagger}$} & \textbf{$f_\psi^{\star}$} \\
\midrule
\textbf{1} & 463 \SE{5} & 351 \SE{14} & 349 \SE{14} & -- & -- & -- \\
\textbf{2} & 457 \SE{5} & 366 \SE{17} & 363 \SE{17} & 165 \SE{5} & 213 \SE{54} & 58 \SE{23} \\
\textbf{4} & 433 \SE{9} & 354 \SE{12} & 351 \SE{13} & 402 \SE{45} & 380 \SE{40} & 158 \SE{59} \\
\textbf{8} & 422 \SE{9} & 365 \SE{18} & 361 \SE{19} & 706 \SE{23} & 606 \SE{31} & 260 \SE{95} \\
\textbf{16} & 402 \SE{8} & 364 \SE{16} & 361 \SE{17} & 1031 \SE{97} & 829 \SE{39} & 279 \SE{106} \\
\textbf{32} & 380 \SE{6} & 351 \SE{16} & 348 \SE{16} & 1270 \SE{39} & 816 \SE{49} & 424 \SE{156} \\
\textbf{64} & 363 \SE{6} & 352 \SE{14} & 350 \SE{14} & 1793 \SE{52} & 1291 \SE{24} & 271 \SE{101} \\
\bottomrule
\end{tabular}
\caption{Per-iteration timing phase breakdown, restricted to the \emph{multiplicative\_coprime} reward.}
\label{tab:timings_multiplicative_coprime}
\end{table}

\begin{table}[h]
\centering
\footnotesize
\setlength{\tabcolsep}{6pt}
\begin{tabular}{rrrrrrr}
\toprule
\textbf{$K$} & \textbf{Sample} & \textbf{Loss} & \textbf{Bwd} & \textbf{Exch.$^\dagger$} & \textbf{$\nabla_\theta^{\ddagger}$} & \textbf{$f_\psi^{\star}$} \\
\midrule
\textbf{1} & 413 \SE{7} & 351 \SE{12} & 350 \SE{12} & -- & -- & -- \\
\textbf{2} & 416 \SE{10} & 363 \SE{15} & 361 \SE{16} & 143 \SE{25} & 123 \SE{17} & 31 \SE{13} \\
\textbf{4} & 404 \SE{10} & 346 \SE{14} & 343 \SE{14} & 352 \SE{95} & 306 \SE{29} & 102 \SE{47} \\
\textbf{8} & 386 \SE{8} & 358 \SE{11} & 354 \SE{12} & 487 \SE{78} & 532 \SE{12} & 223 \SE{83} \\
\textbf{16} & 379 \SE{11} & 353 \SE{10} & 350 \SE{11} & 788 \SE{133} & 697 \SE{25} & 260 \SE{100} \\
\textbf{32} & 362 \SE{6} & 364 \SE{14} & 360 \SE{14} & 967 \SE{144} & 710 \SE{75} & 303 \SE{115} \\
\textbf{64} & 347 \SE{8} & 348 \SE{9} & 344 \SE{9} & 1687 \SE{244} & 1323 \SE{113} & 311 \SE{119} \\
\bottomrule
\end{tabular}
\caption{Per-iteration timing phase breakdown, restricted to the \emph{conditional\_multiscale} reward.}
\label{tab:timings_conditional_multiscale}
\end{table}

\end{document}

%% file: math_commands.tex
\usepackage{amsmath,amsfonts,bm}

\def\eqref#1{equation~\ref{#1}}
\def\1{\bm{1}}

\DeclareMathAlphabet{\mathsfit}{\encodingdefault}{\sfdefault}{m}{sl}
\SetMathAlphabet{\mathsfit}{bold}{\encodingdefault}{\sfdefault}{bx}{n}

%% file: iclr2026_conference.bib
@article{obando2023small,
  title={Small batch deep reinforcement learning},
  author={Obando Ceron, Johan and Bellemare, Marc and Castro, Pablo Samuel},
  journal={Advances in Neural Information Processing Systems},
  volume={36},
  pages={26003--26024},
  year={2023}
}

@inproceedings{
pan2024pretraining,
title={Pre-Training and Fine-Tuning Generative Flow Networks},
author={Ling Pan and Moksh Jain and Kanika Madan and Yoshua Bengio},
booktitle={The Twelfth International Conference on Learning Representations},
year={2024},
url={https://openreview.net/forum?id=ylhiMfpqkm}
}

@article{singla2024sapg,
  title={{SAPG}: Split and Aggregate Policy Gradients},
  author={Singla, Jayesh and Agarwal, Ananye and Pathak, Deepak},
  journal={arXiv preprint arXiv:2407.20230},
  year={2024}
}

@article{bartoldson2025trajectory,
  title={Trajectory balance with asynchrony: Decoupling exploration and learning for fast, scalable {LLM} post-training},
  author={Bartoldson, Brian R and Venkatraman, Siddarth and Diffenderfer, James and Jain, Moksh and Ben-Nun, Tal and Lee, Seanie and Kim, Minsu and Obando-Ceron, Johan and Bengio, Yoshua and Kailkhura, Bhavya},
  journal={arXiv preprint arXiv:2503.18929},
  year={2025}
}

@article{da2024embarrassingly,
  title={Embarrassingly Parallel GFlowNets},
  author={Da Silva, Tiago and Carvalho, Luiz Max and Souza, Amauri and Kaski, Samuel and Mesquita, Diego},
  journal={arXiv preprint arXiv:2406.03288},
  year={2024}
}

@misc{dasilva2024federated,
  title={Federated Contrastive {GFlowNets}},
  author={Silva, Tiago and Souza, Amauri H and Carvalho, Luiz Max and Kaski, Samuel and Mesquita, Diego},
  year={2024},
  howpublished={OpenReview preprint},
  url={https://openreview.net/forum?id=VJDFhkwQg6}
}

@inproceedings{sal2025distributed,
  title={Generalization and Distributed Learning of {GFlowNets}},
  author={Silva, Tiago and Souza, Amauri and Rivasplata, Omar and Garg, Vikas and Kaski, Samuel and Mesquita, Diego},
  booktitle={International Conference on Learning Representations},
  year={2025},
  url={https://openreview.net/forum?id=PJNhZoCjLh}
}

@inproceedings{garipov2023compositional,
  title={Compositional Sculpting of Iterative Generative Processes},
  author={Garipov, Timur and De Peuter, Sebastiaan and Yang, Ge and Garg, Vikas and Kaski, Samuel and Jaakkola, Tommi},
  booktitle={Advances in Neural Information Processing Systems},
  year={2023}
}

@inproceedings{mnih2016a3c,
  title={Asynchronous Methods for Deep Reinforcement Learning},
  author={Mnih, Volodymyr and Badia, Adri{\`a} Puigdom{\`e}nech and Mirza, Mehdi and Graves, Alex and Lillicrap, Timothy and Harley, Tim and Silver, David and Kavukcuoglu, Koray},
  booktitle={International Conference on Machine Learning},
  year={2016}
}

@inproceedings{espeholt2018impala,
  title={{IMPALA}: Scalable Distributed Deep-{RL} with Importance Weighted Actor-Learner Architectures},
  author={Espeholt, Lasse and Soyer, Hubert and Munos, Remi and Simonyan, Karen and Mnih, Volodymyr and Ward, Tom and Doron, Yotam and Firoiu, Vlad and Harley, Tim and Dunning, Iain and others},
  booktitle={International Conference on Machine Learning},
  year={2018}
}

@inproceedings{parkerholder2020dvd,
  title={Effective Diversity in Population Based Reinforcement Learning},
  author={Parker-Holder, Jack and Pacchiano, Aldo and Choromanski, Krzysztof and Roberts, Stephen},
  booktitle={Advances in Neural Information Processing Systems},
  year={2020}
}

@inproceedings{malkin2022trajectory,
  title={Trajectory Balance: Improved Credit Assignment in {GFlowNets}},
  author={Malkin, Nikolay and Jain, Moksh and Bengio, Emmanuel and Sun, Chen and Bengio, Yoshua},
  booktitle={Advances in Neural Information Processing Systems},
  year={2022}
}

@inproceedings{tiapkin2024gfn_entropy,
  title={Generative Flow Networks as Entropy-Regularized {RL}},
  author={Tiapkin, Daniil and Morozov, Nikita and Naumov, Alexey and Vetrov, Dmitry},
  booktitle={International Conference on Artificial Intelligence and Statistics},
  year={2024}
}

@article{rector2023tsgfn,
  title={Thompson Sampling for Improved Exploration in {GFlowNets}},
  author={Rector-Brooks, Jarrid and Madan, Kanika and Jain, Moksh and Korablyov, Maksym and Liu, Cheng-Hao and Chandar, Sarath and Malkin, Nikolay and Bengio, Yoshua},
  journal={arXiv preprint arXiv:2306.17693},
  year={2023}
}

@inproceedings{jain2022biological,
  title={Biological Sequence Design with {GFlowNets}},
  author={Jain, Moksh and Bengio, Emmanuel and Hernandez-Garcia, Alex and Rector-Brooks, Jarrid and Dossou, Bonaventure FP and Ekbote, Chanakya Ajit and Fu, Jie and Zhang, Tianyu and Kilgour, Michael and Zhang, Dinghuai and others},
  booktitle={International Conference on Machine Learning},
  year={2022}
}

@misc{kim2025boosted_gfn,
      title={Boosted {GFlowNets}: Improving Exploration via Sequential Learning},
      author={Dall'Antonia, Pedro and da Silva, Tiago and de Souza, Daniel Augusto and Mattos, C{\'e}sar Lincoln C. and Mesquita, Diego},
      year={2025},
      eprint={2511.09677},
      archivePrefix={arXiv},
      primaryClass={cs.LG},
      url={https://arxiv.org/abs/2511.09677},
}

@inproceedings{ceron2024moe_rl,
  title={Mixtures of Experts Unlock Parameter Scaling for Deep {RL}},
  author={Obando-Ceron, Johan and Sokar, Ghada and Willi, Timon and Lyle, Clare and Farebrother, Jesse and Foerster, Jakob and Dziugaite, Gintare Karolina and Precup, Doina and Castro, Pablo Samuel},
  booktitle={International Conference on Machine Learning},
  year={2024}
}

@article{bengio2021flow,
  title={Flow Network based Generative Models for Non-Iterative Diverse Candidate Generation},
  author={Bengio, Emmanuel and Jain, Moksh and Korablyov, Maksym and Precup, Doina and Bengio, Yoshua},
  journal={Advances in Neural Information Processing Systems},
  volume={34},
  pages={27381--27394},
  year={2021}
}

@article{bengio2023gflownet,
  title={{GFlowNet} Foundations},
  author={Bengio, Yoshua and Lahlou, Salem and Deleu, Tristan and Hu, Edward J and Tiwari, Mo and Bengio, Emmanuel},
  journal={Journal of Machine Learning Research},
  volume={24},
  number={210},
  pages={1--55},
  year={2023}
}

@inproceedings{madan2023learning,
  title={Learning {GFlowNets} from Partial Episodes for Improved Convergence and Stability},
  author={Madan, Kanika and Rector-Brooks, Jarrid and Korablyov, Maksym and Bengio, Emmanuel and Jain, Moksh and Nica, Andrei Cristian and Bosc, Tom and Bengio, Yoshua and Malkin, Nikolay},
  booktitle={International Conference on Machine Learning},
  year={2023}
}

@article{kim2023local,
  title={Local Search {GFlowNets}},
  author={Kim, Minsu and Yun, Taeyoung and Bengio, Emmanuel and Zhang, Dinghuai and Bengio, Yoshua and Ahn, Sungsoo and Park, Jinkyoo},
  journal={arXiv preprint arXiv:2310.02710},
  year={2023}
}

@article{jaderberg2017population,
  title={Population Based Training of Neural Networks},
  author={Jaderberg, Max and Dalibard, Valentin and Osindero, Simon and Czarnecki, Wojciech M and Donahue, Jeff and Razavi, Ali and Vinyals, Oriol and Green, Tim and Dunning, Iain and Simonyan, Karen and Fernando, Chrisantha and Kavukcuoglu, Koray},
  journal={arXiv preprint arXiv:1711.09846},
  year={2017}
}

@article{jacobs1991adaptive,
  title={Adaptive Mixtures of Local Experts},
  author={Jacobs, Robert A and Jordan, Michael I and Nowlan, Steven J and Hinton, Geoffrey E},
  journal={Neural Computation},
  volume={3},
  number={1},
  pages={79--87},
  year={1991}
}

@inproceedings{shazeer2017outrageously,
  title={Outrageously Large Neural Networks: The Sparsely-Gated Mixture-of-Experts Layer},
  author={Shazeer, Noam and Mirhoseini, Azalia and Maziarz, Krzysztof and Davis, Andy and Le, Quoc and Hinton, Geoffrey and Dean, Jeff},
  booktitle={International Conference on Learning Representations},
  year={2017}
}

@inproceedings{horgan2018distributed,
  title={Distributed Prioritized Experience Replay},
  author={Horgan, Dan and Quan, John and Budden, David and Barth-Maron, Gabriel and Hessel, Matteo and van Hasselt, Hado and Silver, David},
  booktitle={International Conference on Learning Representations},
  year={2018}
}

@article{viviano2023torchgfn,
  title={torchgfn: A {PyTorch} {GFlowNet} library},
  author={Viviano, Joseph D. and Younis, Omar G. and Choi, Sanghyeok and Schmidt, Victor and Bengio, Yoshua and Lahlou, Salem},
  journal={arXiv preprint arXiv:2305.14594},
  year={2023}
}

@inproceedings{tiapkin2024generative,
  title={Generative Flow Networks as Entropy-Regularized RL},
  author={Tiapkin, Daniil and Morozov, Nikita and Naumov, Alexey and Vetrov, Dmitry P.},
  booktitle={AISTATS}, series={PMLR}, volume={238}, pages={4213--4221}, year={2024}}

@inproceedings{deleu2024discrete,
  title={Discrete Probabilistic Inference as Control in Multi-path Environments},
  author={Deleu, Tristan and Nouri, Padideh and Malkin, Nikolay and Precup, Doina and Bengio, Yoshua},
  booktitle={UAI}, series={PMLR}, volume={244}, year={2024}}

@inproceedings{zhang2022path,
  title={Path Integral Sampler: A Stochastic Control Approach for Sampling},
  author={Zhang, Qinsheng and Chen, Yongxin}, booktitle={ICLR}, year={2022}}

@inproceedings{vargas2023denoising,
  title={Denoising Diffusion Samplers},
  author={Vargas, Francisco and Grathwohl, Will and Doucet, Arnaud},
  booktitle={ICLR}, year={2023}}

@inproceedings{lahlou2023continuous,
  title={A Theory of Continuous Generative Flow Networks},
  author={Lahlou, Salem and Deleu, Tristan and Lemos, Pablo and Zhang, Dinghuai and Volokhova, Alexandra and Hern{\'a}ndez-Garc{\'i}a, Alex and Ezzine, L{\'e}na N{\'e}hale and Bengio, Yoshua and Malkin, Nikolay},
  booktitle={ICML}, series={PMLR}, year={2023}}

@inproceedings{pan2023gafn,
  title={Generative Augmented Flow Networks},
  author={Pan, Ling and Zhang, Dinghuai and Courville, Aaron and Huang, Longbo and Bengio, Yoshua},
  booktitle={ICLR}, year={2023}}

@inproceedings{shen2023towards,
  title={Towards Understanding and Improving {GFlowNet} Training},
  author={Shen, Max W. and Bengio, Emmanuel and Hajiramezanali, Ehsan and Loukas, Andreas and Cho, Kyunghyun and Biancalani, Tommaso},
  booktitle={ICML}, year={2023}}

@inproceedings{vemgal2023empirical,
  title={An Empirical Study of the Effectiveness of Using a Replay Buffer on Mode Discovery in {GFlowNets}},
  author={Vemgal, Nikhil and Lau, Elaine and Precup, Doina},
  booktitle={ICML 2023 Workshop on Structured Probabilistic Inference \& Generative Modeling (SPIGM)},
  year={2023}, note={arXiv:2307.07674}}

@inproceedings{schaul2016prioritized,
  title={Prioritized Experience Replay},
  author={Schaul, Tom and Quan, John and Antonoglou, Ioannis and Silver, David},
  booktitle={International Conference on Learning Representations (ICLR)},
  year={2016}, note={arXiv:1511.05952}}

@inproceedings{madan2025sibling,
  title={Towards Improving Exploration through Sibling Augmented {GFlowNets}},
  author={Kanika Madan and Alex Lamb and Emmanuel Bengio and Glen Berseth and Yoshua Bengio},
  booktitle={The Thirteenth International Conference on Learning Representations},
  year={2025},
  url={https://openreview.net/forum?id=HH4KWP8RP5}
}

@misc{ikram2024evolution,
      title={Evolution Guided Generative Flow Networks}, 
      author={Zarif Ikram and Ling Pan and Dianbo Liu},
      year={2025},
      eprint={2402.02186},
      archivePrefix={arXiv},
      primaryClass={cs.LG},
      url={https://arxiv.org/abs/2402.02186}, 
}

@article{richardson2026local,
  title={Local Inconsistency Resolution: The Interplay between Attention and Control in Probabilistic Models},
  author={Richardson, Oliver E and Samiei, Mandana and Shakerinava, Mehran and Viviano, Joseph D and Kabid, Abdessamad El and Parviz, Ali and Bengio, Yoshua},
  journal={arXiv preprint arXiv:2604.17140},
  year={2026}
}

@article{younis2023chromax,
  title={ChromaX: a fast and scalable breeding program simulator},
  author={Younis, Omar G and Turchetta, Matteo and Ariza Suarez, Daniel and Yates, Steven and Studer, Bruno and Athanasiadis, Ioannis N and Krause, Andreas and Buhmann, Joachim M and Corinzia, Luca},
  journal={Bioinformatics},
  volume={39},
  number={12},
  pages={btad691},
  year={2023}
}

@article{wang2014wheat,
  title={Characterization of polyploid wheat genomic diversity using a high-density 90{,}000 single nucleotide polymorphism array},
  author={Wang, Shichen and others},
  journal={Plant Biotechnology Journal},
  volume={12},
  number={6},
  pages={787--796},
  year={2014}
}

@article{haldane1919combination,
  title={The combination of linkage values, and the calculation of distances between the loci of linked factors},
  author={Haldane, John Burdon Sanderson},
  journal={Journal of Genetics},
  volume={8},
  number={29},
  pages={299--309},
  year={1919}
}

@article{tester2010breeding,
  title={Breeding technologies to increase crop production in a changing world},
  author={Tester, Mark and Langridge, Peter},
  journal={Science},
  volume={327},
  number={5967},
  pages={818--822},
  year={2010}
}

@article{esquinas2005protecting,
  title={Protecting crop genetic diversity for food security: political, ethical and technical challenges},
  author={Esquinas-Alc{\'a}zar, Jos{\'e}},
  journal={Nature Reviews Genetics},
  volume={6},
  number={12},
  pages={946--953},
  year={2005}
}

@article{donald1968ideotypes,
  title={The breeding of crop ideotypes},
  author={Donald, C. M.},
  journal={Euphytica},
  volume={17},
  number={3},
  pages={385--403},
  year={1968}
}


%% file: iclr2027_conference.bib
@inproceedings{zhang2023robust,
  title={Robust Scheduling with {GFlowNets}},
  author={Zhang, David W. and Rainone, Corrado and Peschl, Markus and Bondesan, Roberto},
  booktitle={International Conference on Learning Representations},
  year={2023}
}
